\documentclass{article}

\usepackage[preprint]{neurips_2026}  %

\usepackage[utf8]{inputenc} %
\usepackage[T1]{fontenc}    %
\usepackage{hyperref}       %
\usepackage{url}            %
\usepackage{booktabs}       %
\usepackage{amsfonts}       %
\usepackage{nicefrac}       %
\usepackage{microtype}      %
\usepackage{xcolor}         %

\usepackage{multirow}
\usepackage[detect-weight, mode=text]{siunitx}
\usepackage[inline]{enumitem}
\usepackage{subcaption}
\usepackage{wrapfig}

\usepackage{array}

\usepackage{bm}
\usepackage{adjustbox}
\usepackage{tcolorbox}
\usepackage{colortbl}
\usepackage{minted}
\usepackage{svg}

\makeatletter
\def\UrlAlphabet{%
      \do\a\do\b\do\c\do\d\do\e\do\f\do\g\do\h\do\i\do\j%
      \do\k\do\l\do\m\do\n\do\o\do\p\do\q\do\r\do\s\do\t%
      \do\u\do\v\do\w\do\x\do\y\do\z\do\A\do\B\do\C\do\D%
      \do\E\do\F\do\G\do\H\do\I\do\J\do\K\do\L\do\M\do\N%
      \do\O\do\P\do\Q\do\R\do\S\do\T\do\U\do\V\do\W\do\X%
      \do\Y\do\Z}
\def\UrlDigits{\do\1\do\2\do\3\do\4\do\5\do\6\do\7\do\8\do\9\do\0}
\g@addto@macro{\UrlBreaks}{\UrlOrds}
\g@addto@macro{\UrlBreaks}{\UrlAlphabet}
\g@addto@macro{\UrlBreaks}{\UrlDigits}
\makeatother

\usepackage{pifont}
\usepackage{marvosym}
\usepackage{fontawesome5}
\usepackage{simpleicons}

\usepackage{tikz}
\usetikzlibrary{shapes.symbols}
\usetikzlibrary{arrows.meta}
\usetikzlibrary{backgrounds}
\usetikzlibrary{positioning, fit, mindmap, trees, calc, tikzmark, shapes}
\usetikzlibrary{shapes.arrows, fadings, automata, tikzmark, decorations.pathreplacing, patterns}
\usetikzlibrary{shapes.geometric}
\usetikzlibrary{intersections}
\usepackage{tkz-euclide}

\usepackage{pgfplots}
\usepgfplotslibrary{groupplots}
\pgfplotsset{compat=1.18}

\usepackage{epigraph}
\definecolor{mine_font}{RGB}{0, 128, 0}
\definecolor{tea_green}{RGB}{214, 234, 193}
\definecolor{hint_green}{RGB}{226,246,209}
\definecolor{Madang}{RGB}{190,235,159}
\definecolor{yellow_green}{RGB}{198,222,119}
\definecolor{link_water}{RGB}{221, 232, 250}
\definecolor{celestial_blue}{RGB}{52, 152, 219}
\definecolor{shakespeare}{RGB}{85, 154, 193}
\definecolor{buttermilk}{RGB}{255,242,174}
\definecolor{chardonnay}{RGB}{250,196,114}
\definecolor{rajah}{RGB}{253,180,98}
\definecolor{fog}{RGB}{213, 193, 234}
\definecolor{melon}{RGB}{254,191,181}
\definecolor{sundown}{RGB}{249, 180, 181}
\definecolor{mona_lisa}{RGB}{246,152,134}
\definecolor{salmon}{RGB}{242,131,107}
\definecolor{blue_x}{RGB}{142, 207, 201}
\definecolor{orange_x}{RGB}{255, 190, 122}

\definecolor{saltpan}{RGB}{238, 243, 232}
\definecolor{aqua_spring}{RGB}{232, 243, 232}
\definecolor{tea_green}{RGB}{214, 234, 193}
\definecolor{Madang}{RGB}{190,235,159}
\definecolor{fringy_flower}{RGB}{194, 234, 193}
\definecolor{aero_blue}{RGB}{193, 234, 213}
\definecolor{pixie_green}{RGB}{183,214,170}
\definecolor{french_pass}{RGB}{195,232,246}
\definecolor{ice_cold}{RGB}{169,232,220}
\definecolor{pale_turquoise}{RGB}{172,240,242}
\definecolor{cruise}{RGB}{179,226,205}
\definecolor{sail}{RGB}{163,205,235}
\definecolor{spindle}{RGB}{179,205,227}
\definecolor{link_water}{RGB}{221, 232, 250}
\definecolor{periwinkle}{RGB}{203,213,232}
\definecolor{zanah}{RGB}{220, 233, 213}
\definecolor{frostee}{RGB}{217, 231, 214}
\definecolor{opal}{RGB}{199, 221, 211}
\definecolor{jet_stream}{RGB}{188, 214, 210}
\definecolor{skeptic}{RGB}{153, 187, 167}
\definecolor{hint_green}{RGB}{226,246,209}
\definecolor{snow_flurry}{RGB}{230,245,201}
\definecolor{surf_crest}{RGB}{205,230,208}
\definecolor{yellow_green}{RGB}{198,222,119}
\definecolor{cream}{RGB}{255,255,204}
\definecolor{pale_prim}{RGB}{255,255,179}
\definecolor{spring_sun}{RGB}{242,243,195}
\definecolor{portafino}{RGB}{245,237,160}
\definecolor{buttermilk}{RGB}{255,242,174}
\definecolor{cream_brulee}{RGB}{255, 229, 151}
\definecolor{dairy_cream}{RGB}{254,226,189}
\definecolor{champagne}{RGB}{254,217,166}
\definecolor{chardonnay}{RGB}{250,196,114}
\definecolor{manhattan}{RGB}{226,180,125}
\definecolor{rajah}{RGB}{253,180,98}
\definecolor{early_dawn}{RGB}{252,243,218}
\definecolor{egg_shell}{RGB}{238, 234, 215}
\definecolor{selago}{RGB}{243, 232, 243}
\definecolor{quartz}{RGB}{219,223,238}
\definecolor{fog}{RGB}{213, 193, 234}
\definecolor{languid_lavender}{RGB}{222,203,228}
\definecolor{watusi}{RGB}{254,221,207}
\definecolor{coral_andy}{RGB}{243,204,205}
\definecolor{cosmos}{RGB}{248,209,210}
\definecolor{melon}{RGB}{254,191,181}
\definecolor{azalea}{RGB}{234, 193, 194}
\definecolor{beauty_bush}{RGB}{235, 185, 179}
\definecolor{sundown}{RGB}{249, 180, 181}
\definecolor{mona_lisa}{RGB}{246,152,134}
\definecolor{salmon}{RGB}{242,131,107}

\definecolor{summer_sky}{RGB}{58, 151, 233}
\definecolor{chateau_green}{RGB}{72, 179, 96}
\definecolor{matisse}{RGB}{25, 104, 167}
\definecolor{allports}{RGB}{31, 106, 125}
\definecolor{sun_shade}{RGB}{255, 144, 68}
\definecolor{flamingo}{RGB}{237, 88, 85}
\definecolor{studio}{RGB}{128, 91, 160}

\definecolor{maya_blue}{RGB}{102, 204, 255}
\definecolor{feijoa}{RGB}{178,223,138}
\definecolor{sushi}{RGB}{117, 168, 47}
\definecolor{norway}{RGB}{158, 194, 132}
\definecolor{japanese_laurel}{RGB}{53, 116, 40}
\definecolor{see_green}{RGB}{161,228,195}
\definecolor{monte_carlo}{RGB}{135,204,194}
\definecolor{granny_smith_apple}{RGB}{150,214,150}
\definecolor{moss_green}{RGB}{170,216,176}
\definecolor{chateau_green}{RGB}{72, 179, 96}
\definecolor{opal}{RGB}{164,207,190}
\definecolor{acapulco}{RGB}{117, 170, 148}
\definecolor{viridian}{RGB}{55, 137, 122}
\definecolor{amazon}{RGB}{56, 123, 84}
\definecolor{asparagus}{RGB}{123, 160, 91}
\definecolor{fruit_salad}{RGB}{91, 160, 94}
\definecolor{puerto_rico}{RGB}{72, 179, 150}
\definecolor{mountain_meadow}{RGB}{0, 163, 136}
\definecolor{matisse}{RGB}{25, 104, 167}
\definecolor{allports}{RGB}{31, 106, 125}
\definecolor{astral}{RGB}{55, 111, 137}
\definecolor{spring_leaves}{RGB}{46, 83, 117}
\definecolor{biscay}{RGB}{44, 62, 80}
\definecolor{midnight}{RGB}{0, 29, 50}
\definecolor{amethyst}{RGB}{153, 102, 204}
\definecolor{studio}{RGB}{128, 91, 160}
\definecolor{tapestry}{RGB}{194, 109, 132}
\definecolor{atomic_tangerine}{RGB}{255, 153, 102}
\definecolor{amber}{RGB}{255, 191, 0}
\definecolor{casablanca}{RGB}{244, 178, 84}
\definecolor{california}{RGB}{233, 140, 58}
\definecolor{tomato}{RGB}{255, 97, 56} 
\definecolor{alizarin}{RGB}{233, 58, 64}

\definecolor{linen}{RGB}{251, 239, 227}
\definecolor{double_pearl_lusta}{RGB}{253, 242, 208}
\definecolor{oasis}{RGB}{253, 242, 208}
\definecolor{milan}{RGB}{255, 254, 169}
\definecolor{texas}{RGB}{245, 232, 123}
\definecolor{maize}{RGB}{249, 212, 156}

\definecolor{turmeric}{RGB}{211, 178, 76}
\definecolor{saffron}{RGB}{249,193,62}
\definecolor{my_sin}{RGB}{255, 176, 59}
\definecolor{tree_poppy}{RGB}{246, 154, 27}
\definecolor{jaffa}{RGB}{240, 131, 58}
\definecolor{crusta}{RGB}{254, 127, 44}
\definecolor{tahiti_gold}{RGB}{223, 102, 36}
\definecolor{outrageous_orange}{RGB}{255, 100, 45}
\definecolor{safety_orange}{RGB}{254, 106, 0}

\definecolor{azalea}{RGB}{251, 196, 196}
\definecolor{oyster_pink}{RGB}{238,206,205} 
\definecolor{coral_candy}{RGB}{242,208,205} 
\definecolor{baby_pink}{RGB}{246, 194, 192}
\definecolor{petite_orchid}{RGB}{223, 157, 155}
\definecolor{apricot}{RGB}{241,140,122}
\definecolor{NY_pink}{RGB}{228,136,113}
\definecolor{carmine_pink}{RGB}{231, 76, 60}
\definecolor{deep_carmine_pink}{RGB}{236, 50, 67}

\definecolor{wewak}{RGB}{244, 143, 150}
\definecolor{light_coral}{RGB}{244, 127, 123}
\definecolor{bittersweet}{RGB}{255,111,105}
\definecolor{carnation}{RGB}{245, 80, 86}
\definecolor{flamingo}{RGB}{237, 88, 85}
\definecolor{sunset_orange}{RGB}{242,89,75}
\definecolor{ku_crimson}{RGB}{243, 0, 25}
\definecolor{amaranth}{RGB}{234,46,73}
\definecolor{valencia}{RGB}{214, 87, 70}
\definecolor{chilean_fire}{RGB}{215, 87, 44}
\definecolor{mexican_red}{RGB}{170, 41, 37}

\definecolor{napa}{RGB}{163, 154, 137}

\definecolor{athens_gray}{RGB}{236, 240, 241}
\definecolor{gallery}{RGB}{240,240,240}
\definecolor{mercury}{RGB}{230,230,230}
\definecolor{platinum}{RGB}{228,228,228}
\definecolor{silver}{RGB}{191,191,191}
\definecolor{aluminum}{RGB}{153,153,153}
\definecolor{ship_gray}{RGB}{77,77,77}
\definecolor{tuatara}{RGB}{67, 67, 67}

\definecolor{malibu}{RGB}{110, 180, 240}
\definecolor{celestial_blue}{RGB}{52, 152, 219}
\definecolor{curious_blue}{RGB}{41, 128, 185}
\definecolor{french_blue}{RGB}{0, 112, 182}
\definecolor{matisse}{RGB}{25, 104, 167}
\definecolor{shakespeare}{RGB}{85, 154, 193}
\definecolor{seagull}{RGB}{128,177,211}
\definecolor{jelly_bean}{RGB}{45, 126, 150}
\definecolor{venice_blue}{RGB}{87, 135, 105}
\definecolor{boston_blue}{RGB}{68, 147, 161}

\definecolor{turquoise}{RGB}{41,217,194}
\definecolor{java}{RGB}{2,190,196}
\definecolor{riptide}{RGB}{141,211,199}
\definecolor{mountain_meadow}{RGB}{0, 163, 136}
\definecolor{free_speech_aquamarine}{RGB}{0, 156, 114}

\definecolor{cosmic_latte}{RGB}{222, 247, 229}
\definecolor{chinook}{RGB}{163, 232, 178}
\definecolor{padua}{RGB}{121, 189, 143}
\definecolor{ocean_green}{RGB}{79, 176, 112}
\definecolor{pastel_green}{RGB}{107, 227, 135}
\definecolor{chateau_green}{RGB}{69, 191, 85}
\definecolor{RoyalBlue}{RGB}{69, 191, 85}
\definecolor{pigment_green}{RGB}{0, 175, 79}
\definecolor{fern}{RGB}{101,197,117}
\definecolor{killarney}{RGB}{56, 113, 66}

\renewcommand{\arraystretch}{0.9}

\newcommand{\unvicon}{\scalebox{0.8}{\tiny\faIcon{university}}}
\newcommand{\shieldicon}{\scalebox{0.8}{\tiny\faIcon{shield-alt}}}

\title{Beyond Reasoning: Reinforcement Learning Unlocks Parametric Knowledge in LLMs}

\author{
 Wanli Yang\textsuperscript{\shieldicon\unvicon} \hspace{0.8em} \textbf{Hongyu Zang} \hspace{0.8em} {\bf Junwei Zhang}\textsuperscript{\unvicon} \\ %
  \textbf{Wenjie Shi} \hspace{0.8em}  \textbf{Du Su}\textsuperscript{\shieldicon} \hspace{0.8em} \textbf{Jingang Wang} \hspace{0.8em} \textbf{Xueqi Cheng}\textsuperscript{\tiny\shieldicon\unvicon} \hspace{0.8em}  \textbf{Fei Sun}\textsuperscript{\shieldicon ~\tiny\faIcon[regular]{envelope}} \\  %
  \textsuperscript{\shieldicon}State Key Laboratory of AI Safety, Institute of Computing Technology, CAS\\
  $\textsuperscript{\unvicon}$University of Chinese Academy of Sciences \\ %
 \texttt{yangwanli24z@ict.ac.cn} \hspace{1em} \texttt{bitwjg@gmail.com} \hspace{1em}
 \textsuperscript{\tiny\faIcon[regular]{envelope}}\texttt{sunfei@ict.ac.cn}
 \vspace{-20pt}
}

\begin{document}

\maketitle

\renewcommand*{\thefootnote}{\tiny\faIcon[regular]{envelope}}
\footnotetext{Corresponding author: Fei Sun (\href{sunfei@ict.ac.cn}{sunfei@ict.ac.cn})}
\renewcommand*{\thefootnote}{\arabic{footnote}}

\begin{abstract}

Reinforcement learning (RL) has achieved remarkable success in LLM reasoning, but whether it can also improve direct recall of parametric knowledge remains an open question. 
We study this question in a controlled zero-shot, one-hop, closed-book QA setting with no chain-of-thought, training only on binary correctness rewards and applying fact-level train-test deduplication to ensure gains reflect improved recall rather than reasoning or memorization. 
Across three model families and multiple factual QA benchmarks, RL yields $\sim$\textbf{27\%} average relative gains, surpassing both training- and inference-time baselines alike. 
Mechanistically, RL primarily redistributes probability mass over existing knowledge rather than acquiring new facts, moving correct answers from the low-probability tail into reliable greedy generations.
Our data-attribution study reveals that the hardest examples are the most informative: those whose answers never appear in 128 pre-RL samples (only $\sim$\textbf{18\%} of training data) drive $\sim$\textbf{83\%} of the gain, since rare correct rollouts still emerge during training and get reinforced. 
Together, these findings broaden the role of RL beyond reasoning, repositioning it as a tool for unlocking rather than acquiring latent parametric knowledge.
  
\end{abstract}

\vspace{-10pt}
\section{Introduction}

Large language models rely on two fundamental capabilities: eliciting parametric knowledge acquired during pre-training and reasoning over such knowledge to produce answers \citep{system1to2}.
Reinforcement learning with verifiable rewards (RLVR) \citep{deepseek2025r1, wen2026reinforcement} has achieved notable success in improving the latter, especially multi-step reasoning in mathematics \citep{yu2025dapo} and coding~\citep{wang2026code}.
However, the former, direct recall of parametric knowledge, is often unreliable and remains largely unexplored: LLMs often ``know more than they express'', producing incorrect answers even when the correct one is encoded in their parameters \citep{orgad2025llms, gekhman2025insideout}.
We therefore ask: \textbf{Beyond complex reasoning, can RL improve the recall of parametric knowledge?}

We show that the answer is \textbf{yes}.
More importantly, RL improves factual recall not by explicit reasoning, but by making latent knowledge more accessible. %
We study this question in a controlled direct-recall setting: \textbf{zero-shot}, \textbf{one-hop closed-book} factual QA, where models are instructed to provide final answers \textbf{without explicit reasoning}.
The RL reward is \textbf{binary and outcome-only}, depending solely on whether the final answer is correct.
We further ensure that held-out test queries share \textbf{no fact-level overlap} with training data, so gains reflect improved recall, not knowledge injected during RL training.
In this setting, RL with binary factual rewards yields substantial improvements across three LLM families and three factual QA benchmarks, with consistent relative gains of $\sim$\textbf{27\% on average} and exceeding \textbf{53\%} on Natural Questions across all three models.
Crucially, these gains transfer robustly across datasets, scale to larger models up to 72B, and persist across RL algorithms, establishing this enhancement as a general property of the RL paradigm.

To understand where these gains come from, we systematically benchmark RL against both training-time and inference-time baselines under identical conditions.
On the training side, supervised fine-tuning (off-policy, positive-only) improves training accuracy without generalizing to held-out queries; DPO \citep{rafailov2023dpo} (off-policy, contrastive) yields limited gains under static preference pairs; and rejection fine-tuning \cite{yuan2023rft} (on-policy, positive-only) achieves smaller and sometimes unstable gains. 
The pattern suggests on-policy exploration and contrastive feedback as the joint source of RL's advantage.
On the inference side, test-time scaling strategies also fall well short of RL: majority voting offers only marginal gains, and chain-of-thought prompting helps inconsistently.
Together, these comparisons establish RL as a distinct paradigm for improving recall of parametric knowledge, one that conventional training- or inference-time methods cannot match.

Having established these gains, we first examine \textit{which failed questions RL repairs, and what distinguishes them from those it does not?}
A natural hypothesis is that RL preferentially recovers factual knowledge the model could already weakly reach, rather than ones that lie entirely outside its reach.
To quantify reachability, we measure pre-RL accessibility as the fraction of the correct answer among 128 stochastic answers drawn from the model before RL.
Our analysis reveals a clear pattern: RL repair rates rise sharply with pre-RL accessibility. 
Partially accessible answers (9–16/128 correct samples) are repaired at $\sim$\textbf{52\%}, and highly accessible answers ($\geq$65/128) at $\sim$\textbf{84\%}. 
Even the hardest cases, whose correct answers are not observed in 128 pre-RL samples, are repaired at 6–13\%, suggesting that some of these facts are encoded but deeply suppressed, not absent.

Beyond \textit{which} questions RL repairs, \textit{how} do these repairs happen in the model's generation distribution? 
When a correct answer becomes top-ranked in the post-RL model, did RL make a previously unreachable fact reachable, or did it move an answer that already existed in the low-probability tail toward the front of the distribution?
To distinguish these cases, we extend the analysis from greedy decoding to pass@$k$ \citep{2024largelanguagemonkeys}, tracking performance as the sampling budget $k$ grows from 1 to 256. 
We find that post-RL accuracy at $k=1$ or $k=2$ often matches what the pre-RL model requires $k=16$ or $k=32$ to achieve, indicating that RL turns a large sampling budget into reliable greedy decoding.
Yet as $k$ grows, the gap closes: under a sufficient sampling budget of $k=256$, the pre-RL model can usually reach the facts that RL unlocks. 
This suggests that RL does not primarily generate new facts; instead, it pulls existing ones from the low-probability tail of the output distribution into reliably top-ranked positions.

Finally, we examine \textit{which training examples drive this redistribution}.
We conduct a controlled data attribution study, stratifying training examples by pre-RL accessibility and training separate RL models on each subset with matched data size.
A natural prediction is that partially accessible examples should dominate: highly accessible facts leave little room to improve, while inaccessible@128 ones appear too sparse to learn from.
Yet the opposite holds.
Although the inaccessible@128 subset accounts for only $\sim$\textbf{18\%} of the full training data, it alone recovers $\sim$\textbf{83\%} of the full-data RL gain; combined with the partially accessible subset, it matches the full-data gain on average.
Tracking the training dynamics reveals why: some of these facts retain a nonzero probability of emerging during repeated rollouts, and once sampled, these rare correct answers are reinforced and progressively amplified over training. 
This reframes what counts as a useful training example for factual RL: the strongest learning signal comes not from facts the model already recalls reliably, but from the low-probability tail of its output distribution.

Our main contributions are summarized as follows:
\begin{itemize}[leftmargin=*, topsep=1pt, itemsep=1pt]
    \item We extend RL beyond reasoning, showing that simple binary rewards substantially improve direct factual recall across diverse models, datasets, and scales.
    \item We show that these gains arise not from injecting new knowledge, but from redistributing probability mass: RL pulls suppressed answers from the low-probability tail into reliably top-ranked positions.
    \item We identify a counterintuitive driver: the strongest training signal comes from facts the pre-RL model rarely recalls, yet RL rollouts can still occasionally elicit.
\end{itemize}

\section{Problem Formulation and Experimental Setup}
\label{sec:background}

In this section, we formulate the problem of RL for direct factual recall, describe our RL training, and detail the experimental setup underlying all subsequent analyses.

\subsection{Problem Formulation: RL for Factual Recall}

To investigate whether RL can improve direct factual recall of parametric knowledge in LLMs, we study a direct factual QA setting: zero-shot, one-hop, closed-book question answering, where the model is instructed to produce a concise final answer without intermediate reasoning steps. 
Formally, given a factual query $q$, the model $\pi_\theta$ generates an answer $a \sim \pi_\theta(\cdot \mid q)$ under a strict non-Chain-of-Thought (non-CoT) constraint (prompt in Appendix~\ref{apd:prompts}), and correctness is determined by a binary indicator $\mathcal{E}(a, a^*) \in \{0, 1\}$.
The non-CoT constraint is designed to minimize confounds from explicit reasoning traces, so that observed improvements are primarily attributable to enhanced factual recall.

\subsection{RL Training}

We adopt Group Relative Policy Optimization (GRPO) \citep{shao2024deepseekmath} as our representative RL algorithm. 
GRPO estimates advantages by contrasting rewards within a rollout group, eliminating the need for a separate value network and making it well-suited for our outcome-based setting. 
Accordingly, we use binary factual correctness as the reward signal, determined via LLM-based semantic verification rather than exact matching, as the latter inherently penalizes semantically correct but differently phrased answers, causing reward sparsity and yielding only marginal gains, as discussed in Section~\ref{sec:discussion}.
For fair evaluation, we maintain a unified hyperparameter configuration across all model-dataset combinations, with full implementation details provided in Appendix~\ref{apd:rl_detail}.

\subsection{Experimental Setup}

\textbf{Models and Datasets.}
We experiment with three open-source instruction-tuned LLMs representing distinct model families: Qwen2.5-7B-Instruct \citep{qwen25technicalreport}, Llama-3.1-8B-Instruct \citep{grattafiori2024llama3}, and OLMo-2-7B-Instruct \citep{olmo20252olmo2furious}. 
For evaluation, we adopt four factual QA benchmarks: Natural Questions (NQ) \citep{kwiatkowski-etal-2019-natural}, TriviaQA \citep{joshi-etal-2017-triviaqa}, PopQA \citep{mallen-etal-2023-popqa}, and SimpleQA \citep{wei2024simpleqa}, spanning a wide range of knowledge types and difficulty levels, from common trivia to long-tail entities and challenging frontier questions.
Following common practice, we partition these datasets into training, validation, and test subsets, subsampling the exceptionally large NQ and TriviaQA training sets to \num{10000} examples. 
Crucially, to ensure that correct answers reflect improved factual recall rather than the mere memorization of training facts, we strictly prevent data contamination by implementing a \textbf{semantic deduplication pipeline}: we identify candidate overlaps via dense embedding similarity and employ LLM-as-a-Judge verification to remove any test query targeting the same underlying fact as a training instance.
Detailed split statistics and deduplication procedures are deferred to Appendix~\ref{apd:data_details}.

\textbf{Generation Strategy.}
For answer generation, we default to greedy decoding for standard evaluations, while for all analytical experiments requiring multiple stochastic samples, we align the sampling hyperparameters with those of the RL training rollouts.

\textbf{LLM-as-a-Judge Verification.}
The scale of our experiments, tens of millions of verification calls across RL training and analytical experiments, necessitates a local open-weight judge to ensure reproducibility and avoid prohibitive API costs. 
To guarantee evaluation quality within these constraints, we select Qwen2.5-72B-Instruct, one of the most capable open-weight models available, as our unified judge for both training rewards and test evaluation. 
Since using the same model for reward assignment and test evaluation may raise reward hacking concerns, we conduct a reliability analysis comparing Qwen against human annotations and frontier closed-source LLMs on 200 sampled outputs spanning pre- and post-RL stages. 
Qwen achieves 92\% overall human agreement, comparable with top-tier proprietary models.
Critically, if reward hacking were occurring, exploiting judge-specific preferences would manifest as degraded human--judge agreement after RL; instead, agreement increases from 89\% to 95\%, and Qwen's false positive rate (answers it accepts that human annotators reject) is exactly 0\% across all 200 samples, explicitly mitigating reward hacking concerns.
Full reliability analysis is provided in Appendix~\ref{apd:llm_judge}.

\section{RL Reliably Improves Direct Factual Recall}
\label{sec:main_exp}

In this section, we systematically evaluate the effectiveness of RL in enhancing direct factual recall.
To understand its underlying mechanisms and examine its generality, we benchmark RL against training and test-time baselines, and further assess its robustness across diverse practical settings.

\subsection{RL's Advantage: On-Policy Exploration Meets Contrastive Feedback}

\begin{table*}[t]
\centering
\captionsetup{belowskip=0pt}
\caption{Main results on four QA benchmarks and three LLMs. We report the accuracy (\%) for different training approaches. The best results are highlighted in \textbf{bold}. TQA, NQ, PQA, and SQA denote TriviaQA, Natural Questions, PopQA, and SimpleQA, respectively.}
\label{tab:main_exp}
\begin{adjustbox}{max width=\textwidth}
\begin{tabular}{l cccc cccc cccc}
\toprule
\multirow{2.5}{*}{\textbf{Method}} & \multicolumn{4}{c}{\textbf{Llama-3.1-8B}} & \multicolumn{4}{c}{\textbf{OLMo-2-7B}} & \multicolumn{4}{c}{\textbf{Qwen2.5-7B}} \\
\cmidrule(lr){2-5} \cmidrule(lr){6-9} \cmidrule(lr){10-13}
& \textbf{TQA} & \textbf{NQ} & \textbf{PQA} & \textbf{SQA} & \textbf{TQA} & \textbf{NQ} & \textbf{PQA} & \textbf{SQA} & \textbf{TQA} & \textbf{NQ} & \textbf{PQA} & \textbf{SQA} \\
\midrule

Base & \num{63.54} & \num{30.15} & \num{23.52} & \num{5.43} & \num{54.86} & \num{22.79} & \num{21.07} & \num{3.56} & \num{49.39} & \num{21.42} & \num{16.96} & \textbf{\num{4.15}} \\
SFT        & \num{62.35}  & \num{29.45}  & \num{29.52}  & \num{5.43}  & \num{57.67}  & \num{25.39}  & \num{25.74}  & \num{3.88}  & \num{51.79}  & \num{23.67}  & \num{22.07}  & \num{3.97}  \\
RFT       & \num{65.21}  & \num{36.70}  & \num{27.75}  & \textbf{\num{5.52}}  & \num{59.63}  & \num{25.33}  & \num{25.93}  & \num{3.65}  & \num{53.65}  & \num{27.03}  & \num{21.16}  & \num{4.01}  \\
DPO        & \num{64.40} & \num{30.48} & \num{24.49} & \num{5.11}  & \num{55.62} & \num{23.06} & \num{21.28} & \num{3.88}  & \num{52.43} & \num{22.94} & \num{17.82} & \textbf{\num{4.15}} \\

\midrule
\rowcolor[HTML]{F2F2F2} %
\textbf{RL} & \textbf{\num{69.89}} & \textbf{\num{46.39}} & \textbf{\num{31.44}} & \num{4.74} & \textbf{\num{65.85}} & \textbf{\num{36.91}} & \textbf{\num{27.48}} & \textbf{\num{4.33}} & \textbf{\num{59.94}} & \textbf{\num{34.12}} & \textbf{\num{22.27}} & \num{3.88} \\

\bottomrule
\end{tabular}
\end{adjustbox}
\end{table*}

To investigate the effectiveness of RL for direct factual recall and understand the contribution of its key components, we compare it against baselines that isolate two individual dimensions: on-policy exploration and contrastive reward signals.  
This yields a strict comparison across four distinct mechanisms: Supervised Fine-Tuning (SFT, off-policy, positive-only), Direct Preference Optimization \citep{rafailov2023dpo} (DPO, off-policy, contrastive), Rejection sampling Fine-Tuning \citep{yuan2023rft} (RFT, on-policy, positive-only), and our RL approach using GRPO (on-policy, contrastive).
For RFT, we implement a standard online iterative pipeline: repeatedly sampling answers from the latest model and fine-tuning on the correct subset.
All methods are evaluated under identical conditions, with full implementation details provided in Appendix~\ref{apd:baseline_imp}.

As shown in Table~\ref{tab:main_exp}, a clear capability hierarchy emerges, with RL delivering the strongest overall performance by a substantial margin.
It consistently achieves the highest accuracy across TriviaQA, NQ, and PopQA, yielding average absolute improvements of about 10\% (around 15\% points on NQ) over the base models.
In contrast, off-policy methods (SFT and DPO) provide limited improvements, indicating that offline optimization is insufficient to improve the underlying recall capability.
While RFT yields occasional improvements over standard SFT via on-policy sampling, its overall performance remains suboptimal compared to RL, highlighting that positive-only signals are insufficient to reliably enhance direct factual recall.
Notably, SimpleQA is the sole exception, where all methods fail to yield meaningful improvements. 
This extreme case suggests that factual RL struggles when the base model rarely produces correct answers, a condition we further analyze in Section~\ref{sec:discussion}.

\begin{wrapfigure}{r}{0.35\linewidth}
    \vspace{-14pt}
    \centering
    \includegraphics[width=\linewidth]{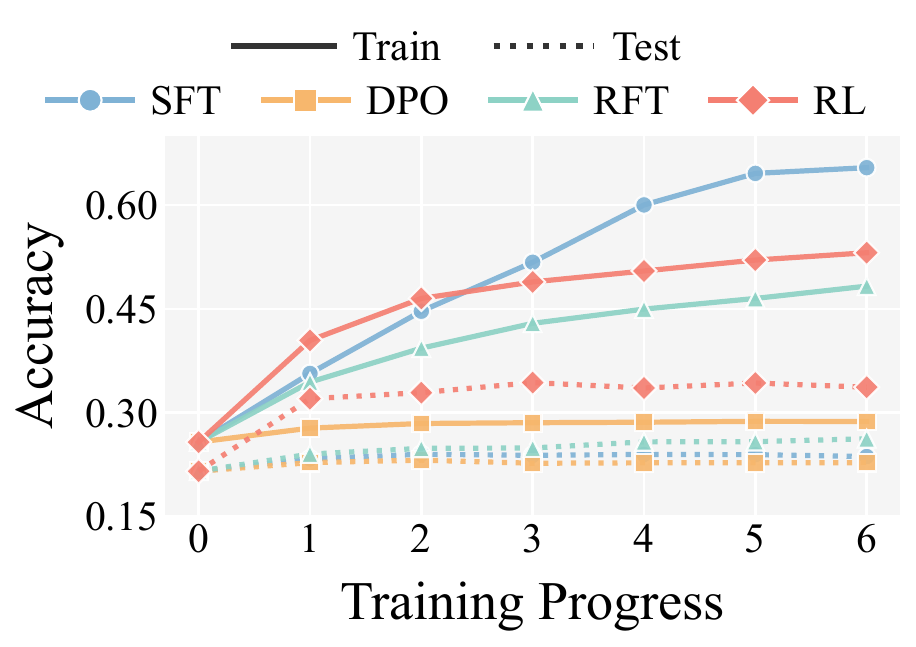}
    \caption{Training dynamics of the four methods on NQ-Qwen.}
    \label{fig:training_dynamics_qwen}
    \vspace{-10pt}
\end{wrapfigure}

To further understand why RL outperforms the baselines, Figure~\ref{fig:training_dynamics_qwen} compares their training dynamics on NQ across normalized training progress, using Qwen as a representative example, with complete results across all three models presented in Appendix~\ref{apd:training_dynamics}.
The baselines exhibit distinct failure modes.
SFT rapidly overfits the training data without generalizing; DPO leaves both curves flat under static preference pairs; and while RFT's on-policy sampling yields minor test-time gains, its lack of negative signals limits its effectiveness. 
Conversely, driven by active exploration and advantage-based reward signals, RL effectively optimizes the general factual recall capability, yielding uniquely large, sustained improvements on the test set.

\subsection{RL Achieves What Inference-Time Scaling Cannot}

\begin{figure}
    \centering
    \includegraphics[width=\linewidth]{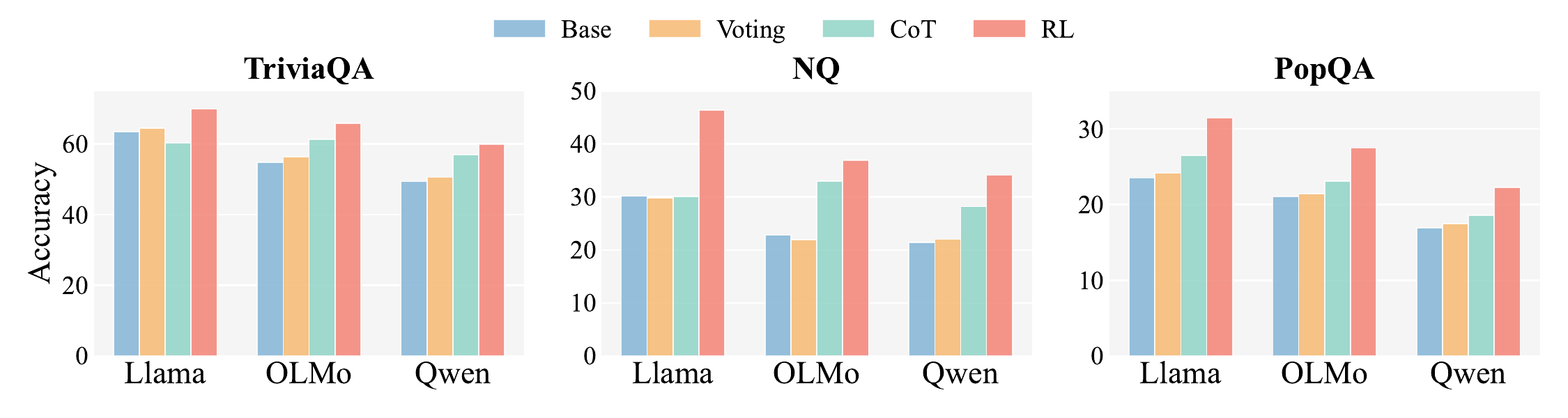}
    \captionsetup{skip=2pt, belowskip=0pt}
    \caption{Comparison between test-time scaling strategies and RL across various datasets and LLMs. 
    \textbf{Voting} denotes majority voting over 32 independently sampled answers from the base model.}
    \label{fig:inference_baseline}
\end{figure}

Beyond training-time optimization, a prevailing paradigm for more effectively leveraging parametric knowledge is test-time scaling \citep{snell2025scaling, muennighoff-etal-2025-s1}.
To determine whether scaling inference compute can replicate RL's gains in factual recall, we compare RL against two representative test-time strategies applied to the base models: majority voting \citep{wang2023selfconsistency} and chain-of-thought (CoT) prompting \citep{wei2022chain}.
For majority voting, we return the most frequent normalized response from 32 independent direct-answer generations, a practical budget, with alternative sample sizes discussed in Appendix~\ref{apd:vote_budget}.
For CoT, we prompt for step-by-step reasoning before producing a final answer via single greedy decoding (prompt in Appendix~\ref{apd:prompts}).

As depicted in Figure~\ref{fig:inference_baseline}, majority voting yields only marginal gains over the base model, indicating that while multiple sampling trials can occasionally capture correct facts, they fail to reliably promote the correct answer over incorrect candidates when the truth is not the dominant mode.
CoT proves more effective, consistent with prior evidence that explicit reasoning can partially unlock parametric knowledge \citep{gekhman2026thinking}. 
However, its improvements are inconsistent across models and datasets, and remain substantially below the gains achieved by RL.
In contrast, RL delivers large and consistent improvements across all nine model-dataset combinations, confirming that the benefits of RL cannot be replicated by test-time scaling alone.

\subsection{RL Gains Are Robust Across Datasets, Scales, Architecture, and Algorithms}

Having established RL's unique advantage over alternative approaches, we further examine whether this superiority reflects a general property of the paradigm rather than an artifact of specific configurations. Specifically, we evaluate the robustness of our findings along the following three dimensions.

\begin{wraptable}{r}{0.35\textwidth}
    \vspace{-10pt} %
    \centering
    \caption{Accuracy on NQ across different RL algorithms.} %
    \label{tab:rl_algo}
    \begin{adjustbox}{max width=0.35\textwidth}
    \begin{tabular}{lrrr}
    \toprule
    \textbf{Method} & \textbf{Llama} & \textbf{OLMo} & \textbf{Qwen} \\
    \midrule
    Pre-RL & 30.15 & 22.79 & 21.42 \\
    GRPO & 46.39 & 36.91 & 34.12 \\
    PPO & 46.42 & 37.12 & 32.48 \\
    \bottomrule
    \end{tabular}
    \end{adjustbox}
\end{wraptable}

\textbf{RL algorithms.}
We further investigate whether the observed gains are specific to GRPO or stem from the broader RL paradigm. 
As shown in Table~\ref{tab:rl_algo}, substituting GRPO with Proximal Policy Optimization (PPO) \citep{schulman2017ppo} under identical reward and hyperparameter configurations yields comparable performance across all evaluated models. 
This consistency confirms that the improvement is not an artifact of a specific algorithmic implementation, but rather reflects a fundamental advantage of RL.

\begin{wrapfigure}{r}{0.35\linewidth}
    \vspace{-12pt}
    \centering
    \includegraphics[width=\linewidth]{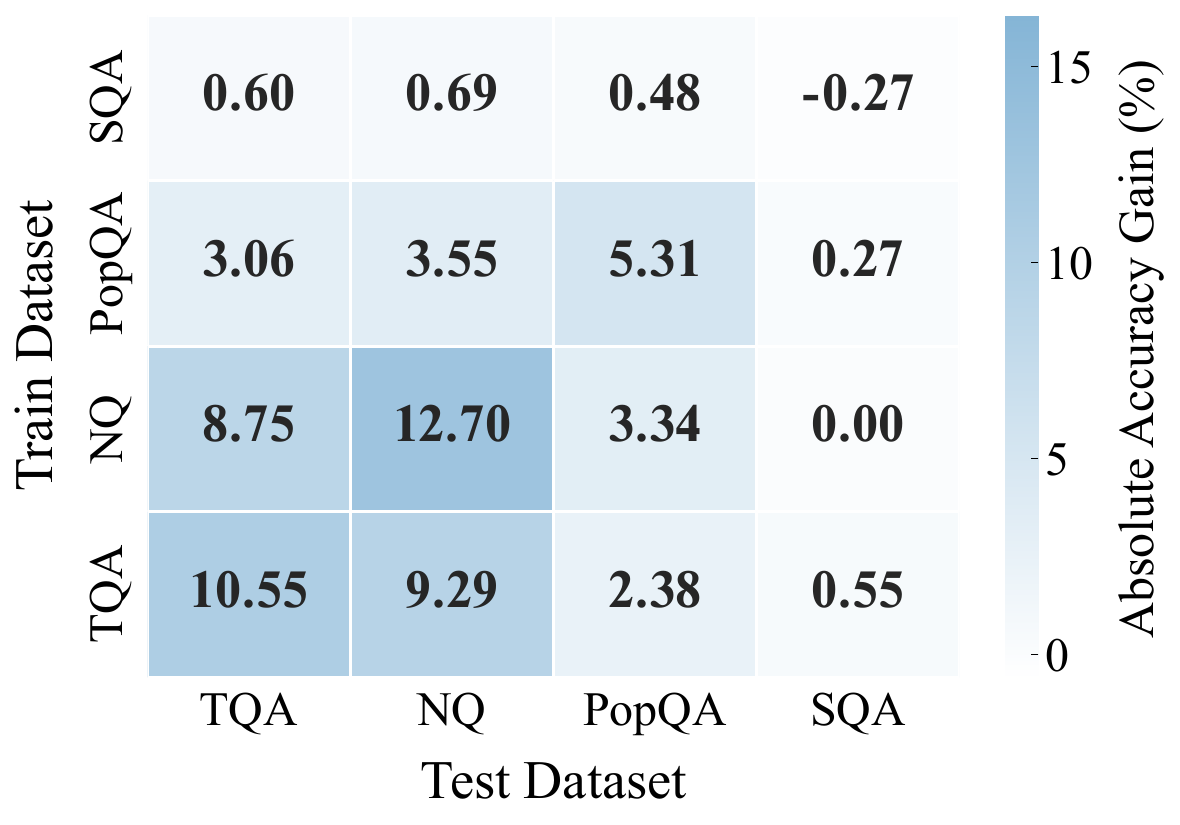}
    \captionsetup{skip=2pt, belowskip=-10pt}
    \caption{Cross-dataset acc. gain.}
    \label{fig:cross-dataset_qwen}
\end{wrapfigure}

\textbf{Cross-dataset transfer.} 
Beyond the in-domain setting, we further examine whether the improvement in factual recall transfers across datasets by training on one QA dataset and evaluating on another. 
We apply the same fact-level deduplication procedure to remove overlapping facts between the source training set and the target test set. 
This setting poses a significant challenge, as the source and target datasets differ substantially in knowledge domains and query styles.
However, as shown in Figure~\ref{fig:cross-dataset_qwen} (using Qwen as a representative case, with full results across all models deferred to Appendix~\ref{apd:cross_dataset}), a highly consistent pattern emerges: excluding combinations involving the exceptionally challenging SimpleQA, RL training yields notable accuracy gains across all cross-dataset pairs. 
These results indicate that the recall improvement is not limited to in-domain evaluation, but transfers robustly to out-of-distribution factual queries.

\begin{table*}[t]
    \centering
    \caption{Effectiveness of RL across different model sizes and architectures on NQ dataset.} 
    \label{tab:model_scale}
    \begin{adjustbox}{max width=\textwidth}
    \begin{tabular}{l r r r r r}
    \toprule
    \textbf{NQ} & \textbf{Qwen2.5-7B} & \textbf{Qwen2.5-14B} & \textbf{Qwen2.5-32B} & \textbf{Qwen2.5-72B} & \textbf{Qwen3-30B-A3B} \\

    \midrule
    
    Pre-RL & \num{21.42} & \num{25.24} & \num{27.94} & \num{34.64} & \num{31.42} \\
    Post-RL & \num{34.12} & \num{41.88} & \num{42.06} & \num{49.61} & \num{43.58} \\

    \bottomrule 
    \end{tabular}
    \end{adjustbox}
\end{table*}

\textbf{Model scale and architecture.}
To verify whether the effectiveness of RL for direct factual recall extends to the larger, more capable models typically deployed in practice, we expand our evaluation to larger dense models (up to 72B in the Qwen2.5 series \citep{qwen25technicalreport}) and a Mixture-of-Experts (MoE) architecture (Qwen3-30B-A3B-Instruct \citep{yang2025qwen3technicalreport}) on the NQ dataset. 
As presented in Table~\ref{tab:model_scale}, RL training consistently yields substantial absolute accuracy gains of approximately 15\%, indicating that its benefits are not restricted to a specific parameter scale or dense architecture.

Collectively, these results establish both the efficacy of RL in enhancing direct factual recall and its robustness under dataset transfer, model scaling, and RL algorithm variants.

\section{RL Reshapes Access to Latent Parametric Knowledge}

\label{sec:repair}

While the main results establish that RL yields significant improvements in direct factual recall, aggregate accuracy alone does not reveal the source of these gains.
In this section, we examine the underlying effect of RL on factual recall: \textit{which initially failed queries are repaired, and how the accessibility of correct answers changes after RL}.

\subsection{RL Preferentially Repairs More Accessible Facts}

\begin{figure}
    \centering
    \includegraphics[width=\linewidth]{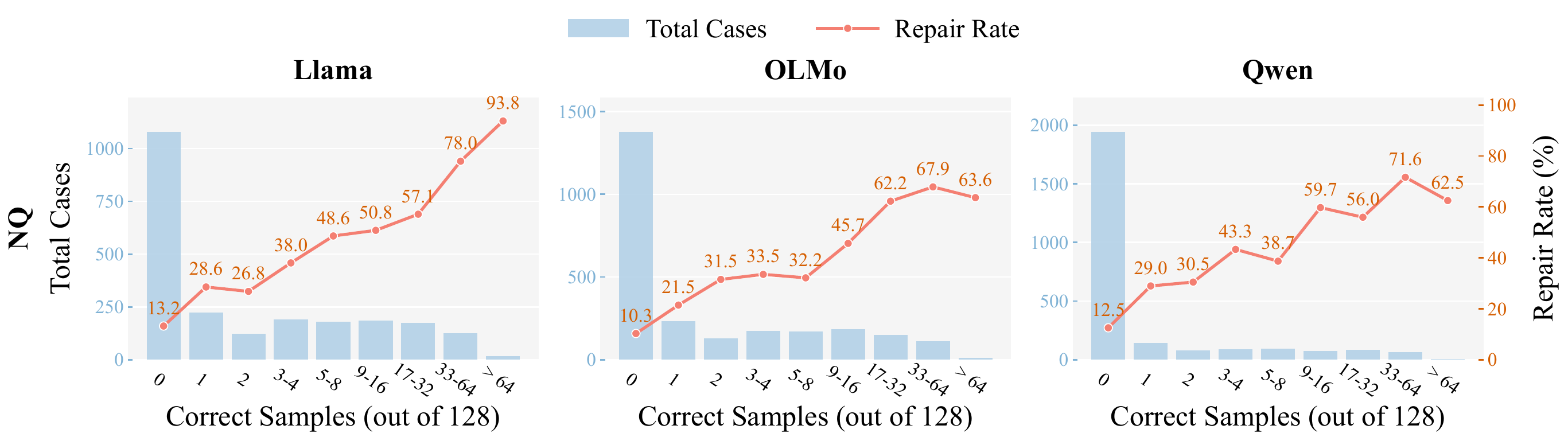}
    \caption{Post-RL repair rates for initially failed test queries, stratified by the pre-RL accessibility of the correct answer.
    Accessibility is measured as the number of times the correct answer appears in 128 independent stochastic samples from the base model.
    Bars show the fraction of queries in each accessibility bin that the post-RL model answers correctly under greedy decoding.}
    \label{fig:repair_rate_avg-acc}
\end{figure}

A natural question is whether RL repairs failed queries indiscriminately, or preferentially recovers a specific subset.
To investigate this, we focus on test queries where the base model fails under greedy decoding. 
Even among these consistently failed queries, the underlying probability of generating the correct answer varies significantly.  
We quantify this probability via \textbf{pre-RL accessibility}, the frequency of the correct answer across 128 independent stochastic samples drawn using the same hyperparameters as the RL rollout phase. 
This metric is not intended to prove whether a fact is stored or absent in the model, but to provide a practical proxy for how readily a fact can be elicited from the output distribution, avoiding the complexity of aggregating token-level logits across diverse answer phrasings. %
Given the long-tail distribution of these frequencies, we categorize these queries into discrete, logarithmically spaced bins based on their correct sample counts: 0, 1, 2, [3, 4], [5, 8], [9, 16], [17, 32], [33, 64], and $\ge 65$.
Finally, we define the \textbf{repair rate} as the fraction of queries within each bin that the post-RL model successfully answers via greedy decoding.

\textbf{Post-RL repair rates are strongly stratified by pre-RL accessibility.}
Figure~\ref{fig:repair_rate_avg-acc} reveals a remarkably consistent pattern across all models on NQ: the probability that RL repairs a failed query rises sharply with the query's pre-RL accessibility, with results on other datasets detailed in Appendix~\ref{apd:extended_repair}.
For instance, queries whose correct answers appear only 9-16 times out of 128 samples (roughly 10\% initial probability) achieve repair rates around 52\%, while highly accessible facts (e.g., $\ge 64$ samples) consistently exceed 62\% and peak at over 93\%.
Since this analysis is conducted on held-out test queries, this amplification cannot be attributed to the model simply sampling and reinforcing these specific facts during training. 
Instead, this pattern suggests that RL broadly strengthens the model's recall capability: facts closer to the surface are more readily brought into greedy-decodable range.

\textbf{RL repairs even facts that are invisible under finite pre-RL sampling.}
A particularly striking result emerges in the zero-accessibility bin: even for queries where the correct answer never appears across all 128 pre-RL samples, RL successfully elevates the fact to the greedy decoding output at a rate of 6\%–13\%.
Since these queries are held out from RL training, the model never receives direct supervision or reward on these specific facts. Their recovery therefore suggests that RL amplifies deeply suppressed parametric signals, rather than injecting new factual knowledge.
Accordingly, zero observed hits under a finite sampling budget should not be taken as proof that the corresponding fact is absent from the model. Rather, they indicate that the fact is not expressed under the base decoding distribution, even though it may still be recoverable after RL.

\subsection{RL Pulls Correct Answers from the Tail Toward the Top}

\begin{figure}
    \centering
    \includegraphics[width=\linewidth]{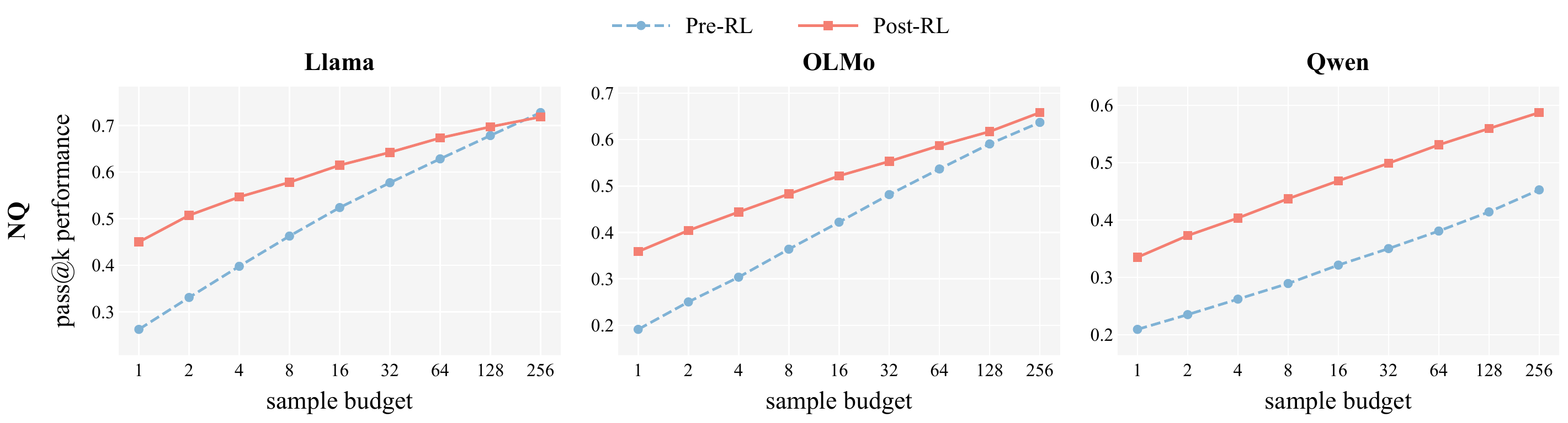}
    \captionsetup{skip=4pt, belowskip=-10pt}
    \caption{Pass@$k$ scaling curves for pre-RL and post-RL models on the NQ dataset.}
    \label{fig:passk_comp}
\end{figure}

While the repair rate analysis provides a query-level perspective on greedy decoding, this top-1 evaluation overlooks underlying improvements where a correct answer becomes significantly easier to sample despite not reaching rank one. 
To further reveal the impact of RL on the model's global recall behavior, we extend our evaluation to stochastic sampling via pass@$k$, which counts a query as solved if the correct answer appears within the first $k$ independent samples.
By scaling this budget up to $k=256$ over the full test set, pass@$k$ provides a direct diagnostic of how rapidly correct answers emerge before and after RL.

\textbf{RL shifts correct answers toward lower-budget recall.}
As shown in Figure~\ref{fig:passk_comp} (NQ; full results in Appendix~\ref{apd:extended_passk}), post-RL models consistently outperform their pre-RL counterparts at low and medium sampling budgets ($k \le 64$). 
Notably, post-RL performance at $k \in \{1, 2\}$ matches what pre-RL models require substantially larger budgets, such as $k \in \{16, 32\}$, to achieve. 
This indicates that RL's gains are not confined to crossing the greedy-decoding threshold: correct answers also become easier to elicit under stochastic sampling, implying a broader increase in factual accessibility.
As the sampling budget increases towards $k=256$, the gap between pre-RL and post-RL curves narrows substantially, suggesting that many correct answers were already reachable under sufficiently large pre-RL sampling budgets but are shifted by RL into a lower-budget recall regime. 

In summary, the query-level repair analysis and the pass@$k$ scaling results demonstrate a consistent conclusion: RL preferentially repairs facts based on their initial accessibility, systematically promoting recoverable but suppressed latent knowledge into more accessible recall regimes.

\section{Lower-Accessibility Examples Contribute More to RL Gains}
\label{sec:attribution}

Having analyzed the structure of RL's gains at test time, we turn to the corresponding training-time question: \textit{which training examples drive these gains?}
To investigate this, we conduct a controlled training data attribution study on TriviaQA and NQ across three LLMs.

Given that pre-RL accessibility strongly correlates with post-RL repair behavior, we employ this metric to partition the training data and identify which subsets provide effective learning signals. 
Based on the number of correct responses ($c$) across 128 pre-RL samples, we partition the training data into four regions: inaccessible@128 ($c=0$), near-inaccessible ($1 \leq c \leq 2$), partially accessible ($3 \leq c \leq 64$), and highly accessible ($c \geq 65$). 
We isolate the training contributions by evaluating three primary subsets (inaccessible@128, partially accessible, highly accessible) and their pairwise combinations. 
We exclude near-inaccessible subset, as 1-2 correct responses out of 128 are difficult to distinguish from random sampling noise. 
To ensure fair comparisons, we keep all RL hyperparameters fixed and balance data sizes within each comparison group: primary subsets and pairwise combinations are separately downsampled to the same size.
Finally, we measure the efficacy of each subset by the fraction of the full-data RL gain it recovers: $(\text{Acc}_{\text{subset}} - \text{Acc}_{\text{base}}) / (\text{Acc}_{\text{full}} - \text{Acc}_{\text{base}})$.

\begin{wrapfigure}[12]{r}{0.4\columnwidth}
    \vspace{-12pt}
    \centering
    \includegraphics[width=\linewidth]{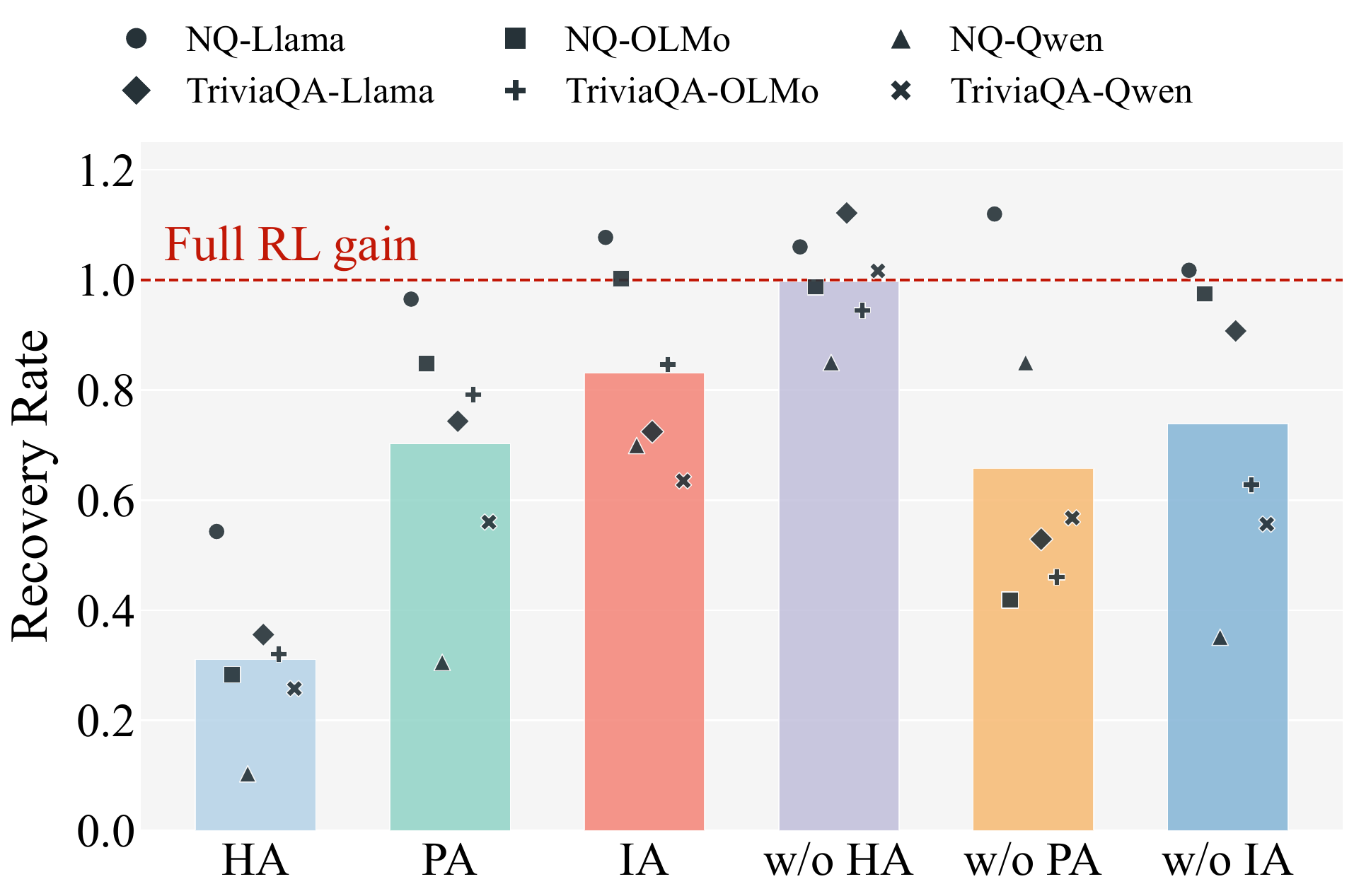}
    \captionsetup{skip=3pt}
    \caption{Recovery of full-data RL gains by training on different data subsets.}
    \label{fig:train_attribution}
\end{wrapfigure}

\paragraph{Inaccessible@128 facts provide the strongest single-subset signal.}
For outcome-based RL, a natural intuition is that partially accessible examples should drive the largest training gains: highly accessible facts offer limited learning signals since reliable successes leave diminished advantages, while inaccessible@128 facts appear to provide prohibitively sparse rewards.
However, Figure~\ref{fig:train_attribution} reveals a counterintuitive pattern.
Among three primary subsets—Highly Accessible (HA), Partially Accessible (PA), and Inaccessible@128 (IA)—the IA subset emerges as the strongest contributor, recovering 83\% of full-data RL gain on average and notably outperforming partially accessible (70\%) and highly accessible (31\%) subsets.

\begin{wrapfigure}{r}{0.3\columnwidth}
    \vspace{-10pt}
    \centering
    \includegraphics[width=\linewidth]{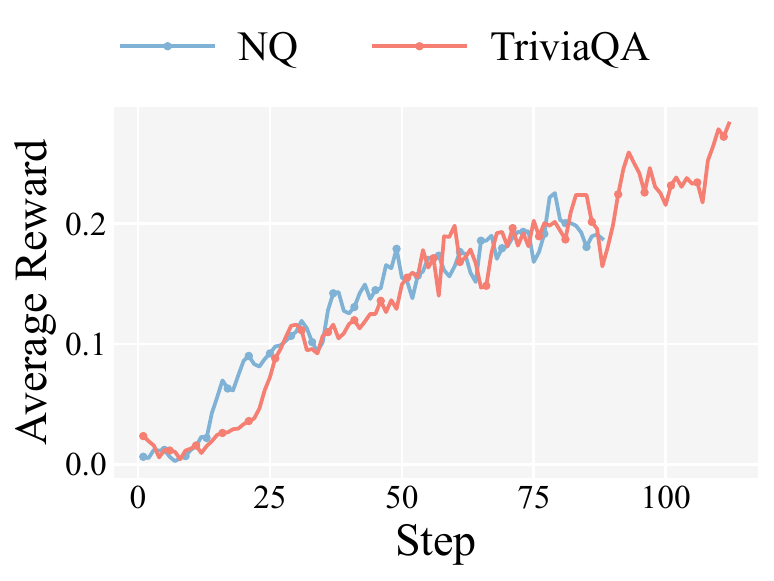}
    \captionsetup{skip=3pt}
    \caption{Reward dynamics of inaccessible@128 training.} %
    \label{fig:reward_dynamics_unknown_qwen}
\end{wrapfigure}

\paragraph{Repeated rollouts capture and reinforce suppressed knowledge.}
To investigate how examples with a $0/128$ pre-RL success rate drive such gains, Figure~\ref{fig:reward_dynamics_unknown_qwen} tracks their reward dynamics, using Qwen as a representative case with full results in Appendix~\ref{apd:reward_dynamics}.
The average reward starts extremely low, confirming that correct responses are indeed rare, but it rises steadily throughout training.
This indicates that finite-sample inaccessibility does not imply the absence of parametric knowledge. 
Rather, some of these facts appear to be suppressed: while they remain hidden within 128 pre-RL samples, they are occasionally generated during repeated RL rollouts. 
Once these rare correct responses occur, their signals are progressively reinforced. 
Crucially, optimizing these highly challenging examples provides the most potent learning signal, driving the most pronounced improvements in overall factual recall.

\paragraph{Low-accessibility data nearly recover the full-data RL gains.}
Furthermore, the pairwise comparisons in Figure~\ref{fig:train_attribution} reinforce the importance of low-accessibility data: combining the partially accessible and inaccessible@128 subsets yields the highest overall gain, recovering the full-data RL gain on average (1.00). 
This indicates that training on low-accessibility data encourages the model to improve its capacity for extracting deeply suppressed knowledge, thereby boosting overall factual recall. 
Ultimately, these results reveal a fundamental property of factual RL: the most valuable training signal lies not in facts the model already recalls reliably, but in latent knowledge it struggles or entirely fails to surface prior to RL.

\section{Discussion: When Factual RL Works, and What It Means for RL}
\label{sec:discussion}

Our analyses reveal a consistent picture of how RL reshapes direct factual recall.
In this section, we further discuss two broader questions: under what conditions does factual RL succeed, and what does this imply for RL more generally?

\begin{wraptable}{r}{0.38\textwidth}
    \vspace{-10pt} %
    \centering
    \renewcommand{\arraystretch}{1}
    \caption{Accuracy on NQ under different reward metrics.} %
    
    \label{tab:ablation_reward}
    \begin{adjustbox}{max width=0.38\textwidth}
    \begin{tabular}{lrrr}
    \toprule
    \textbf{Reward} & \textbf{Llama} & \textbf{OLMo} & \textbf{Qwen} \\
    \midrule
    Pre-RL & 30.15 & 22.79 & 21.42 \\
    EM & 32.12 & 24.15  & 25.12 \\
    LLM-judge & 46.39 & 36.91 & 34.12 \\
    \bottomrule
    \end{tabular}
    \end{adjustbox}
\end{wraptable}

\paragraph{Conditions for successful factual RL.}
Our results confirm two common failure modes for RL: reward saturation on highly accessible facts yields limited group-relative advantages, whereas near-zero initial accuracy (e.g., SimpleQA) leaves the model struggling with extremely sparse rewards \citep{zhang2025interplay}.
Table~\ref{tab:ablation_reward} further corroborates the reward sparsity constraint: replacing LLM-based semantic verification with a strict exact-match metric induces reward sparsity, similarly collapsing RL gains.
This establishes a practical guideline: factual RL requires datasets with non-trivial initial accuracy. 
Counterintuitively, within such datasets, the most valuable training signals come from inaccessible@128 examples, since correct responses for these queries may still surface during repeated rollouts, allowing rare positive rewards to be captured and progressively amplified.
Consequently, data curation for factual RL should prioritize these low-accessibility examples over those of intermediate difficulty.

\paragraph{Broadening the RL paradigm beyond reasoning.}
Outcome-based RL is conventionally viewed as an optimizer for reasoning trajectories to solve complex tasks.
Our work broadens this paradigm, demonstrating that RL can directly optimize factual recall without CoT, yielding substantial held-out gains that neither supervised fine-tuning nor test-time scaling can replicate. 
Crucially, the underlying process is not knowledge injection, but probability redistribution: RL systematically shifts suppressed correct answers from the low-probability tail toward reliable generation. 
This positions RL as a powerful latent knowledge optimizer that enhances the utilization of existing parametric knowledge, achieved neither through reasoning nor through knowledge injection.

\section{Related Work}

\textbf{Parametric Knowledge Recall in LLMs.}
LLMs can output only part of their parametric knowledge \citep{orgad2025llms, gekhman2025insideout}.
Behavioral studies find direct prompting to provide only a lower-bound estimate of encoded knowledge \citep{jiang-etal-2020-know, gekhman-etal-2024-fine, elazar-etal-2021-measuring}, while representation-level studies further show that internal states can contain more hidden factual signals \citep{burns2023discovering, azaria-mitchell-2023-internal, li2023iti}. 
To improve knowledge recall, prior work has largely relied on inference-time elicitation. \textit{Query-level methods} view recall as an addressability problem, exploring prompts to route models toward the target knowledge \citep{shin-etal-2020-autoprompt, zhong-etal-2021-factual}. \textit{Generation-level methods} view recall as a context-construction problem, either using chain-of-thought reasoning \citep{gekhman2026thinking, wei2022chain, wang2023selfconsistency} or asking models to output relevant knowledge before answering \citep{liu-etal-2022-generated, sun2023recitationaugmented}.
In this work, we move from inference-time elicitation to training-time optimization, asking whether knowledge recall can be improved using reinforcement learning.

\textbf{RL for Reasoning and Knowledge Recall.}
RL has largely improved LLMs' \textit{reasoning capability}, benefiting tasks such as mathematics \citep{shao2024deepseekmath, deepseek2025r1} and code generation \citep{le2022coderl, liu2023rltf}. 
Some analyses attribute this gain to on-policy exploration of different rollouts \citep{chu2025sft, wang2026reinforcement, trung-etal-2024-reft}, while others debate the boundary of this effect \citep{yue2025does, wen2026reinforcement, liu2026prorl}.
More recent work also applies RL for \textit{knowledge recall}.
Some methods train models to deliberate over knowledge \citep{ma2026improving}, others target factual precision within reasoning traces \citep{ren2025knowrl, li2025reasoning} or long-form generation \citep{chen2025learning}. 
These studies still focus on reasoning-mediated knowledge use and do not answer whether RL can optimize direct knowledge recall. 
We therefore explore this question by studying single-hop closed-book QA under a strict non-CoT setup.

\textbf{RL in Knowledge-Intensive QA.}
Early work applies RL to \textit{open-domain} QA to optimize interaction with external evidence, either to train retrieval-side policies \citep{wang2018r, buck2018ask, wang-etal-2024-retrieve} or to improve answer generation \citep{nakano2021webgpt, menick2022teaching}.
More relevant to our study, other work applies RL to \textit{closed-book QA}, either to improve instruction following \citep{ziegler2019fine, ouyang2022training, rafailov2023dpo} or to mitigate hallucination \citep{wei2025truthrl, li2025reasoning}.
Such objectives improve QA reliability but conflate factual recall with other goals.
In contrast, we directly study RL for knowledge recall itself.

\section{Conclusion}
\label{sec:conclusion}

This work characterizes the effect of RL on factual recall, beyond its familiar role in reasoning, and the resulting picture changes how we should interpret factual failures in LLMs. 
A wrong answer under greedy decoding is not necessarily evidence that a fact is absent, but can instead reflect an accessibility failure: the correct answer is stranded in the low-probability tail of the output distribution. 
RL addresses this without injecting new knowledge. 
By reinforcing the model's own rare correct rollouts, it turns latent parametric knowledge into reliable direct recall.
This broadens RL from an optimizer of reasoning trajectories to an optimizer of access to memory, and clarifies when factual RL should work: the target knowledge can be hard to recall, but must remain occasionally discoverable. 
Ultimately, what a model knows and what it can express are not the same, and our results show that RL can narrow this accessibility gap.

\bibliographystyle{plainnat}
\bibliography{arxiv}

\clearpage

\appendix

\section{Prompts for Generation}
\label{apd:prompts}

We employ two prompt templates throughout the experiments.
For direct factual QA, the primary setting of this work, models are instructed to produce a single concise entity without any intermediate reasoning, as shown in Figure~\ref{fig:prompt_direct}.
For the chain-of-thought (CoT) baseline, models are required to reason step-by-step before providing a structured final answer, as presented in Figure~\ref{fig:prompt_cot}.

\renewcommand{\fcolorbox}[4][]{#4}
\begin{figure*}[t]
    \begin{tcolorbox}[
        right=5pt, left=5pt, top=5pt, bottom=5pt,
        toptitle=1mm, bottomtitle=1mm,
        colback=white,
        coltitle=white,
        colbacktitle=matisse,
        colframe=matisse,
        title=Prompt for Direct Factual QA, center title]
    \begin{minted}[fontsize=\small,autogobble,numberblanklines=false,breaklines]{markdown}
System: You are a question answering model, answer the following question with a single entity, as concise as possible.

User: Question: {question}
Answer:
    \end{minted}
    \end{tcolorbox}
    \caption{Prompt for direct factual QA. Models are instructed to produce a single concise entity without any intermediate reasoning.}
    \label{fig:prompt_direct}
\end{figure*}

\renewcommand{\fcolorbox}[4][]{#4}
\begin{figure*}[t]
    \begin{tcolorbox}[
        right=5pt, left=5pt, top=5pt, bottom=5pt,
        toptitle=1mm, bottomtitle=1mm,
        colback=white,
        coltitle=white,
        colbacktitle=matisse,
        colframe=matisse,
        title=Prompt for Factual QA with CoT, center title]
    \begin{minted}[fontsize=\small,autogobble,numberblanklines=false,breaklines]{markdown}
System: Answer the following question.

Before giving the final answer, think step by step to recall the relevant factual knowledge carefully and avoid jumping to a premature answer.

Then output your final answer in the following format:
Final answer: <answer>

User: Question: {question}
    \end{minted}
    \end{tcolorbox}
    \caption{Prompt for factual QA with CoT. Models are instructed to reason step by step before producing a final answer in a structured format.}
    \label{fig:prompt_cot}
\end{figure*}

\section{Details of RL Training}
\label{apd:rl_detail}

\textbf{Algorithm.}
We adopt Group Relative Policy Optimization (GRPO) \citep{shao2024deepseekmath} as our representative outcome-based RL algorithm. 
Unlike standard PPO \citep{schulman2017ppo}, GRPO eliminates the memory overhead of a separate value network by estimating advantages through relative reward comparisons within a group of $n$ rollouts sampled for the same query. 
Specifically, the advantage for each rollout is computed by standardizing its reward against the group's mean and standard deviation, ensuring stable policy updates based purely on relative group performance.

\textbf{Reward function.}
We employ a binary reward formulation based on factual correctness: a response receives a reward of $1$ if it is deemed correct, and $0$ otherwise. 
Considering that correct answers may be expressed in various valid phrasings, correctness is evaluated by an LLM-as-a-Judge using semantic verification, as detailed in Appendix~\ref{apd:llm_judge}, rather than exact string matching.

\textbf{Hyperparameters.}
To demonstrate the robustness and generalizability of RL-induced factual knowledge gains, we adopt a unified hyperparameter configuration across all models and datasets, strictly avoiding dataset-specific tuning. 
All RL training is implemented using the VeRL framework~\citep{sheng2024verl}. 
Specifically, we maintain a constant learning rate of $1 \times 10^{-6}$, a global batch size of $128$, and train for $8$ epochs. The policy objective utilizes a KL divergence regularization coefficient of $\beta = 0.001$ and a PPO clip ratio of $\epsilon = 0.2$. 
For rollout generation, we utilize vLLM~\citep{Kwon2023vllm} with a temperature of $T = 1.0$, top-$k = -1$, top-$p = 1.0$, and a group size of $n = 5$ samples per query.

\textbf{Computational Resources.}
All RL training experiments were conducted on a single compute node equipped with 8 NVIDIA A100 (80GB) GPUs. 
The total training time across our experimental pipeline amounted to approximately 80 wall-clock hours.

\section{Data Preparation: Splits and Deduplication}
\label{apd:data_details}

In this section, we provide detailed information regarding our dataset partitions and the rigorous deduplication pipeline employed to ensure the rigor of our evaluation.

\subsection{Data Splits and Statistics}
\label{apd:data_splits}

\begin{wraptable}{r}{0.45\textwidth}
    \centering
    \caption{Statistics of dataset.} %
    \label{tab:data_split}
    \begin{adjustbox}{max width=0.45\textwidth}
    \begin{tabular}{lrrrr}
        \toprule
        \textbf{Split} & \textbf{TQA} & \textbf{NQ} & \textbf{PQA} & \textbf{SQA} \\
        \midrule
        Train                 & \num{10000} & \num{10000} & \num{5000} & \num{2000} \\
        Validation            & \num{512}   & \num{512}   & \num{256}  & \num{128}  \\
        Test                  & \num{11313} & \num{3610}  & \num{9011} & \num{2198} \\
        Test (filtered)       & \num{10802} & \num{3300}  & \num{8689} & \num{2192} \\
        \bottomrule
    \end{tabular}
    \end{adjustbox}
\end{wraptable}

We evaluate our approach across four factual QA benchmarks: Natural Questions (NQ), TriviaQA, PopQA, and SimpleQA. The data partitioning strategies for these datasets are detailed as follows.
For NQ and TriviaQA, whose original training splits each exceed \num{80000} examples, we randomly sample \num{10000} examples for training and reserve a small held-out portion for validation.
Following common practice, we adopt their validation splits as test sets, since NQ lacks an official test set and the test annotations of TriviaQA are not publicly available.
For PopQA and SimpleQA, which provide only a single evaluation set, we randomly partition them into training, validation, and test subsets.
The final statistics for all dataset splits are summarized in Table~\ref{tab:data_split}.

\subsection{Data Deduplication Pipeline}
\label{apd:data_dedup}

To ensure that improvements in held-out performance reflect genuine enhancements in factual recall rather than the mere memorization of training instances, we implement a two-stage semantic deduplication pipeline. 
This process explicitly removes any test query that targets the same underlying fact as any training instance.

\textbf{Stage 1: Embedding-based candidate retrieval.}
We encode all questions across both the training and test sets using \texttt{bge-large-en-v1.5} \citep{bge_embedding}.
For each test sample, we identify the top-10 most similar training queries based on cosine similarity, retaining only pairs with a similarity score above $0.8$ as candidates for further verification.

\textbf{Stage 2: LLM-based semantic verification.}
Each identified candidate pair is passed to Qwen2.5-72B-Instruct. Using a structured prompt (detailed in Figure~\ref{fig:prompt_dedup}), the model is instructed to determine whether both questions target the exact same underlying fact. 
A test sample is removed if any of its candidate training pairs are judged as semantically equivalent.
Specifically, this equivalence is rigorously defined: questions sharing the same subject entity but asking about different relations or sharing the same target answer but originating from different entities, are explicitly treated as non-overlapping distinct facts.

\section{LLM-as-a-Judge: Prompt and Reliability Analysis}
\label{apd:llm_judge}

\textbf{Prompt.}
Adapted from the evaluation protocol of SimpleQA \citep{wei2024simpleqa}, Figure~\ref{fig:prompt_judge} presents the complete prompt template used for our LLM-as-a-Judge evaluation. 
Given a question, a gold target answer, and a predicted response, the model is instructed to assess semantic equivalence and return a strict binary score of 1.0 (correct) or 0.0 (incorrect). 
To ensure robust evaluation, the prompt incorporates few-shot demonstrations of typical correct and incorrect scenarios, explicitly covering valid phrasing variations and incomplete or contradictory responses.

\begin{table*}[t]
    \centering
    \caption{
        Detailed reliability analysis of LLM judges against human annotations ($n=100$ per stage).
        Acc: accuracy with human annotations as ground truth;
        TP: judge and human both correct;
        TN: judge and human both incorrect;
        FP: judge accepts answer human rejects;
        FN: judge rejects answer human accepts.
    }
    \begin{adjustbox}{max width=\textwidth}
    \begin{tabular}{ll rrrrrr}
        \toprule
        & 
        & \textbf{DeepSeek-V3.2}
        & \textbf{Qwen2.5-72B}
        & \textbf{GPT-5}
        & \textbf{GPT-5.4}
        & \textbf{Gemini 2.5 Flash}
        & \textbf{Gemini 3 Flash} \\
        \midrule

        \multirow{5}{*}{\rotatebox[origin=c]{90}{\textbf{Pre-RL}}}
            & Acc & 87.0 & 89.0 & 91.0 & 94.0 & 85.0 & 95.0 \\
            & TP        & 43   & 45   & 47   & 50   & 41   & 51   \\
            & TN        & 44   & 44   & 44   & 44   & 44   & 44   \\
            & FP        & \textbf{0} & \textbf{0} & \textbf{0} & \textbf{0} & \textbf{0} & \textbf{0} \\
            & FN        & 13   & 11   & 9    & 6    & 15   & 5    \\
        \midrule

        \multirow{5}{*}{\rotatebox[origin=c]{90}{\textbf{Post-RL}}}
            & Acc & 93.0 & 95.0 & 91.0 & 95.0 & 83.0 & 96.0 \\
            & TP        & 57   & 59   & 55   & 59   & 47   & 60   \\
            & TN        & 36   & 36   & 36   & 36   & 36   & 36   \\
            & FP        & \textbf{0} & \textbf{0} & \textbf{0} & \textbf{0} & \textbf{0} & \textbf{0} \\
            & FN        & 7    & 5    & 9    & 5    & 17   & 4    \\
        \bottomrule
    \end{tabular}
    \end{adjustbox}
    \label{tab:judge}
\end{table*}

\textbf{Reliability analysis.}
To validate the reliability of Qwen2.5-72B-Instruct as our unified judge and mitigate potential reward-hacking concerns, we conduct a human agreement analysis alongside comparisons with frontier models, including DeepSeek-V3.2, GPT-5, GPT-5.4, Gemini 2.5 Flash, and Gemini 3 Flash. 
We randomly sample 200 model outputs covering pre- and post-RL stages, two datasets (NQ and TriviaQA), and two model families (Qwen and Llama), with 25 samples per cell.
Table~\ref{tab:judge} reports the agreement rate of each judge with human annotations.

Overall, Qwen2.5-72B demonstrates exceptional reliability. 
With an average agreement rate of 92.00\%, it explicitly outperforms GPT-5 (91.00\%) and Gemini 2.5 Flash (84.00\%) on this straightforward factual verification task, while remaining highly competitive with the latest flagship APIs (GPT-5.4 and Gemini 3 Flash). 
Furthermore, for most judges, agreement rates increase from the pre-RL to the post-RL stage, suggesting that RL-optimized models produce more definitive outputs that are inherently easier to assess. 
Particularly, Qwen2.5-72B achieves 95.00\% agreement for post-RL outputs, strictly on par with the best-performing models. 
Crucially, FP = 0 across all judges and both stages, meaning no judge ever accepts an answer that human annotators consider incorrect; accuracy differences are driven entirely by FN.
This directly confirms that the observed post-RL gains are not artifacts of reward hacking, and substantiates the use of Qwen2.5-72B as a reliable judge for both training and evaluation.

\section{Implementation of Baselines}
\label{apd:baseline_imp}

\textbf{Direct Preference Optimization (DPO).}
We construct preference data for DPO \citep{rafailov2023dpo} directly from the QA dataset.
For each training question, the ground-truth answer serves as the chosen response. 
To construct a representative rejected response, we independently sample 16 candidate outputs from the base model and select the most frequent incorrect answer, acting as a hard negative. 
All DPO training is executed utilizing the Llama-Factory framework \citep{zheng-etal-2024-llamafactory}.

\textbf{Rejection Fine-Tuning (RFT).}
We implement RFT \citep{yuan2023rft} through an iterative, on-policy data generation and training pipeline. 
At each iteration, the current model independently generates 5 candidate responses per question, aligning with the rollout group size used in our RL training. 
If at least one response is verified as correct, we randomly sample one correct output to serve as the fine-tuning target; queries yielding zero correct answers are discarded for this iteration. 
The model is then fine-tuned for a single epoch on these curated targets using Llama-Factory. 
This cycle is repeated for up to 15 iterations, with early stopping triggered if held-out validation accuracy fails to improve for 5 consecutive iterations.

\section{Extended Training Dynamics Across All Models}
\label{apd:training_dynamics}

\begin{figure}
    \centering
    \includegraphics[width=\linewidth]{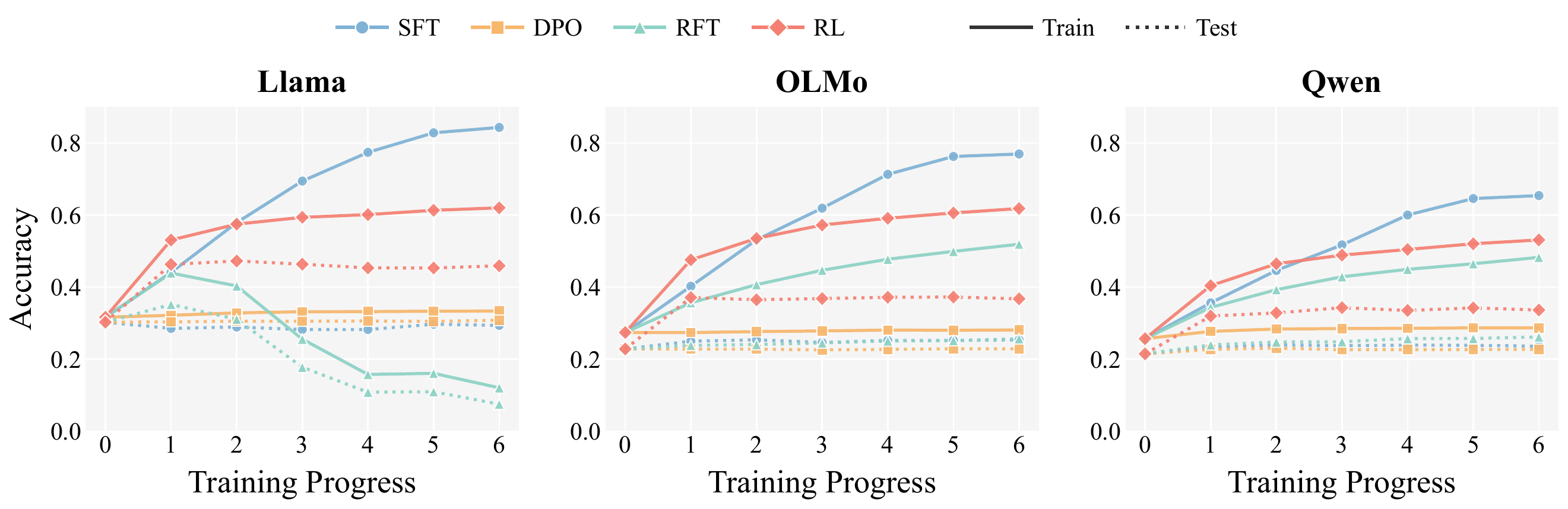}
    \caption{Training dynamics of the four methods on the NQ benchmark across three LLMs. Solid and dashed lines denote training and test accuracy, respectively. To account for varying total training steps, the x-axis represents six equally spaced fractions of each method's respective training progress.}
    \label{fig:app_training_dynamics_all}
\end{figure}

In the main text, we utilized Qwen as a representative example to illustrate the training dynamics of different methods. 
In this section, we provide the complete training dynamics across all three evaluated model families: OLMo, Llama, and Qwen on NQ dataset in Figure~\ref{fig:app_training_dynamics_all}.

The extended results corroborate the distinct failure modes of the baseline methods discussed in the main text. 
Specifically, SFT consistently demonstrates rapid overfitting across all models, achieving high training accuracy while yielding negligible improvements on the held-out test set. 
Similarly, DPO struggles to optimize generative factual recall, resulting in consistently flat curves across all models.
Notably, the full results highlight the inherent instability of RFT. 
While RFT achieves minor test-time gains on Qwen and OLMo, its reliance on positive-only on-policy sampling without contrastive negative signals proves highly brittle. 
This is most evident on Llama, where RFT induces severe optimization instability, leading to a sharp performance collapse after an initial peak.
In contrast, RL, driven by active exploration and advantage-based reward signals, consistently produces robust, uniquely large, and sustained test-set improvements across all three diverse model architectures. 
This cross-model consistency confirms that the factual recall enhancements driven by RL are a general property of the paradigm rather than an artifact of a specific model family.

\section{Discussion of the Majority Voting Budget}
\label{apd:vote_budget}

\begin{figure}[t]
    \centering
    \includegraphics[width=\linewidth]{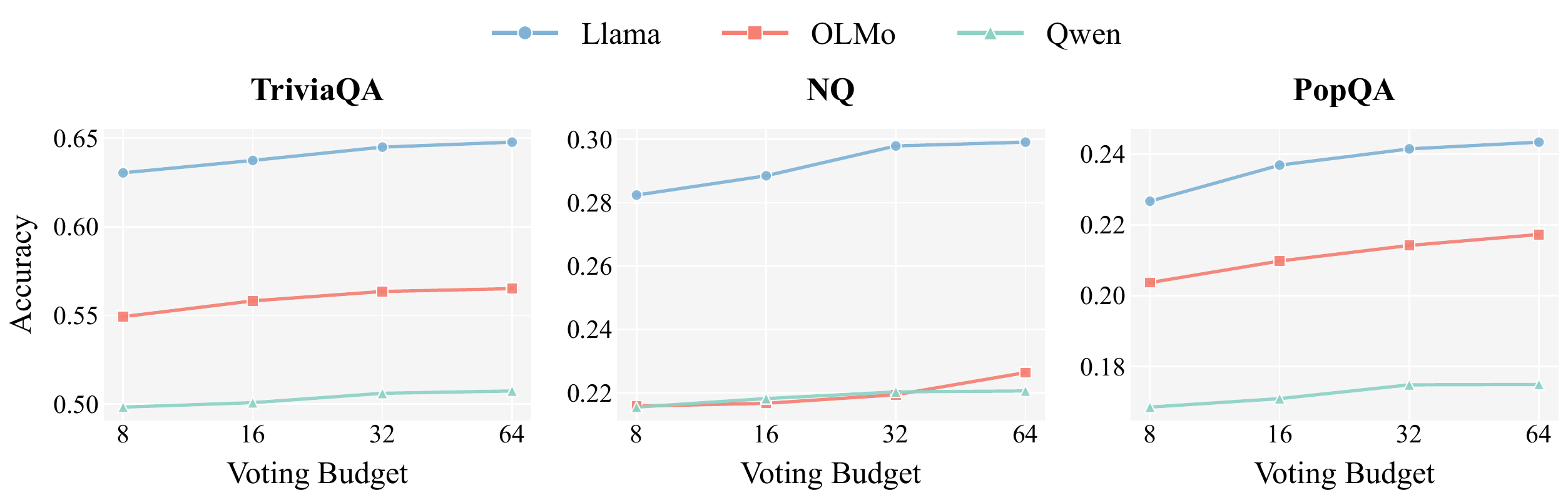}
    \caption{Accuracy of majority voting at different sampling budgets ($k \in \{8, 16, 32, 64\}$).}
    \label{fig:vote_budget}
\end{figure}

To assess whether majority voting offers a complementary inference-time benefit, we apply it to the direct-answer setting using the same sampling configuration as our RL rollouts. 
We evaluate voting budgets of $8$, $16$, $32$, and $64$ samples. 
As shown in Figure~\ref{fig:vote_budget}, scaling the voting budget has no meaningful effect on accuracy across all tested settings. 
This outcome is expected: majority voting selects the most frequently generated answer, which approximates the mode of the output distribution. 
When the model's greedy output is incorrect, simply increasing the voting budget cannot recover the correct fact, as it remains trapped in the low-probability tail. 
This corroborates our finding: the gains from RL stem from a redistribution of probability mass, fundamentally improving factual recall in a way that cannot be replicated by mere inference-time scaling.

\section{Extended Cross-Dataset Transfer Results}
\label{apd:cross_dataset}

\begin{figure}
    \centering
    \includegraphics[width=\linewidth]{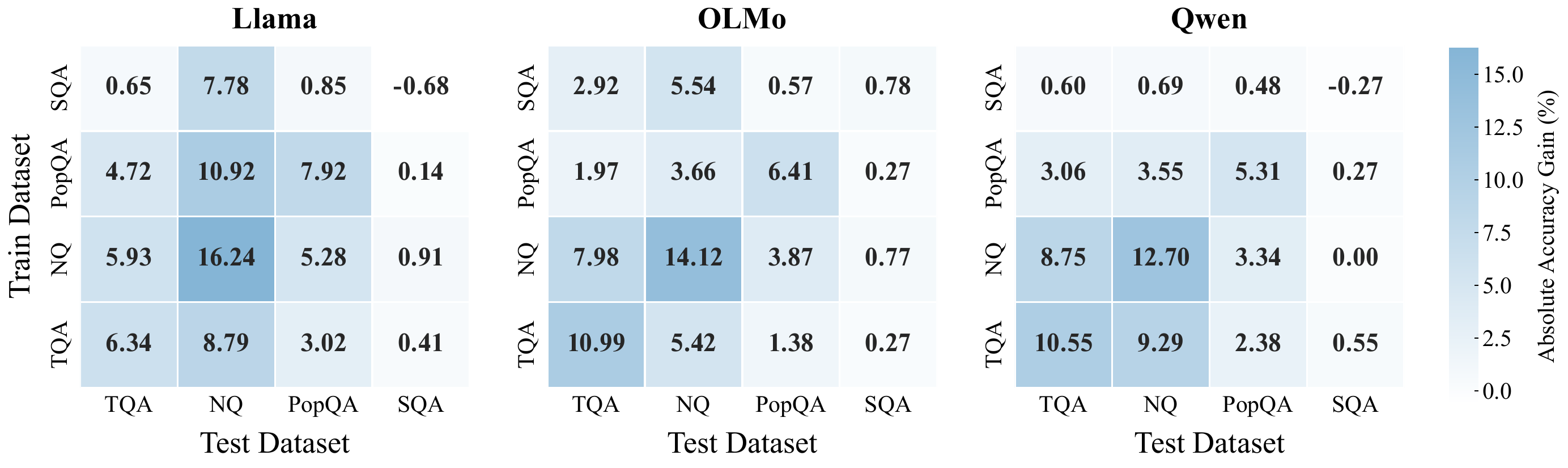}
    \caption{Cross-dataset accuracy gains achieved by RL over original models.}
    \label{fig:app_cross_dataset_all}
\end{figure}

In the main text, we evaluated the cross-dataset transferability of RL-enhanced factual recall using Qwen as a representative case. 
In this section, we present the comprehensive cross-dataset evaluation across all three model architectures: OLMo, Llama, and Qwen. 
Figure~\ref{fig:app_cross_dataset_all} illustrates the zero-shot transfer performance when models are trained on a source QA dataset and evaluated on an out-of-distribution target dataset, following our rigorous fact-level deduplication.

The extended results confirm that the robust transferability observed on Qwen is a universal characteristic across different LLM families. 
Despite substantial variations in knowledge domains and query formatting between the source and target datasets, RL training consistently yields significant relative accuracy improvements across almost all cross-dataset pairs for both Llama and OLMo. 
The only consistent exception across all architectures is the transfer to SimpleQA. 
As an exceptionally challenging benchmark, its difficulty largely exceeds the inherent factual capacity of the evaluated base models, naturally leading to limited transfer gains. 
Overall, these comprehensive results substantiate that RL fundamentally optimizes a generalizable factual recall mechanism, rather than merely overfitting to the superficial stylistic features of the training domain.

\section{Extended Results of Post-RL Repair Rates}
\label{apd:extended_repair}

\begin{figure}
    \centering
    \includegraphics[width=\linewidth]{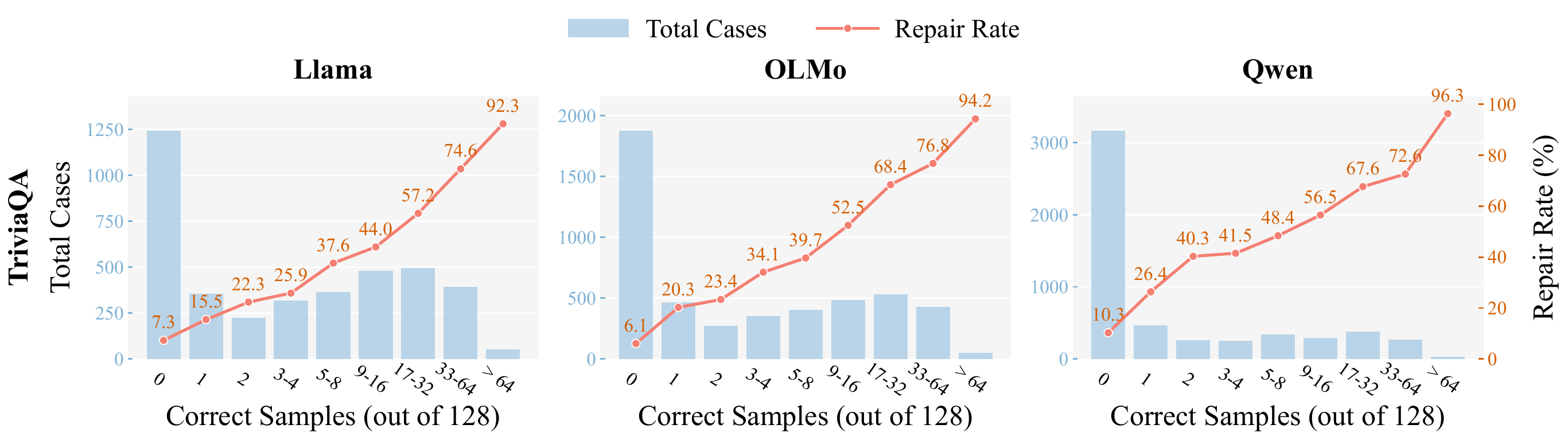}
    \caption{Post-RL repair rates for initially failed test queries on TriviaQA.}
    \label{fig:app_repair_rate_all}
\end{figure}

In the main text, we analyzed the relationship between pre-RL accessibility and post-RL repair rates using the NQ dataset as a representative case.
In this section, we present the comprehensive repair rate analysis extended to another mainstream dataset, TriviaQA.
Figure~\ref{fig:app_repair_rate_all} illustrates these post-RL repair rates, notably stratified by pre-RL accessibility.

The extended results demonstrate that the core phenomena observed on NQ are highly consistent across different knowledge domains.
Specifically, two key patterns universally hold:
First, the post-RL repair rate exhibits a strong, monotonic positive correlation with pre-RL accessibility across all datasets and model architectures.
Second, the phenomenon of zero-accessibility recovery consistently emerges: RL successfully elevates deeply suppressed facts that never appear within the initial finite sampling budget to the greedy decoding output.
These comprehensive results empirically confirm that the mechanisms by which RL amplifies suppressed parametric knowledge are generalizable and not an artifact of the NQ dataset.

\section{Extended Results of Pass@$k$ Scaling}
\label{apd:extended_passk}

\begin{figure}
    \centering
    \includegraphics[width=\linewidth]{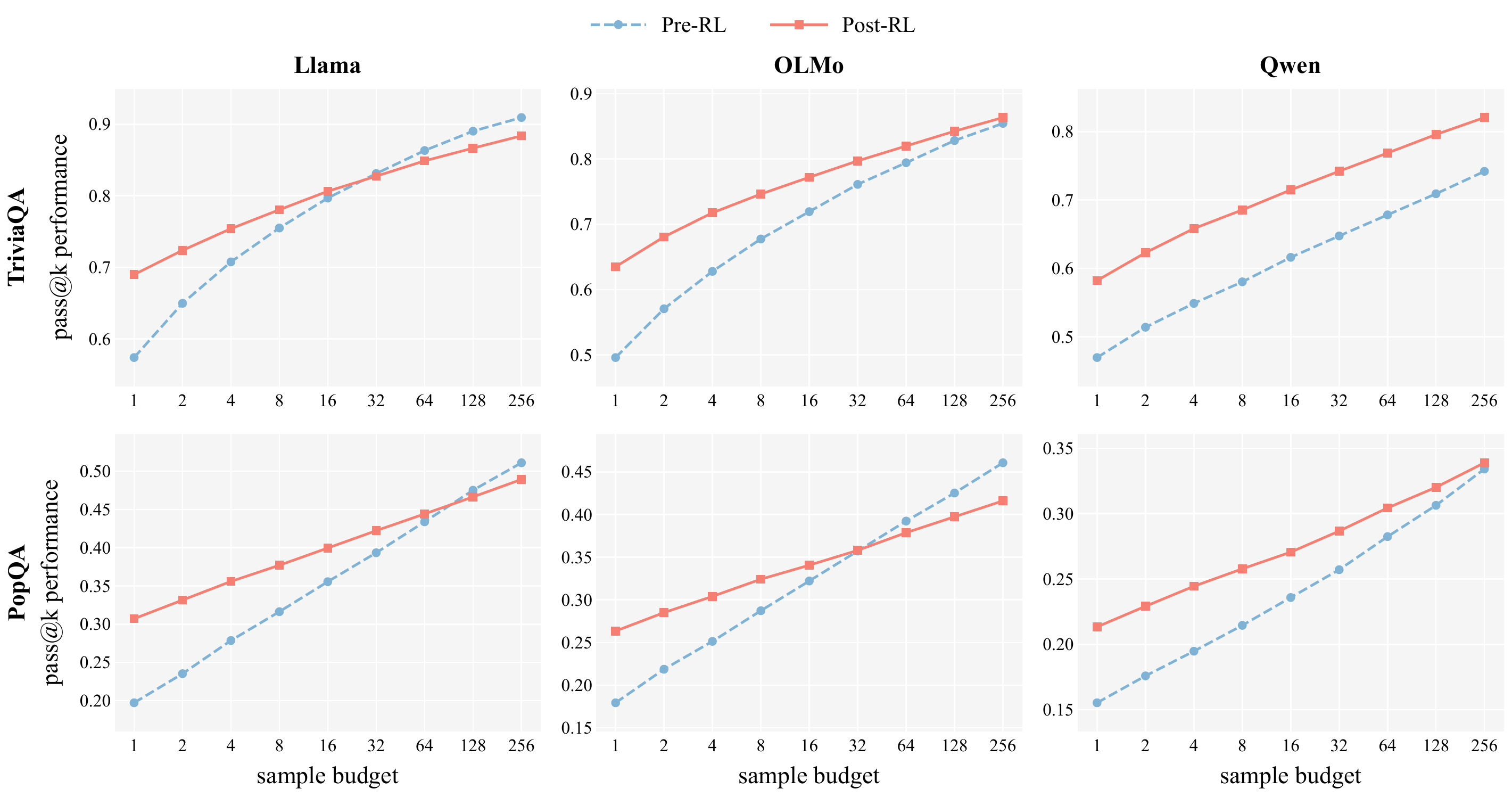}
    \caption{Pass@$k$ scaling curves for pre-RL and post-RL models on TriviaQA and PopQA.}
    \label{fig:app_passk_all}
\end{figure}

In the main text, we analyzed the global recall behavior of pre-RL and post-RL models via pass@$k$ scaling, using the NQ dataset as a representative example. 
In this section, we present the comprehensive pass@$k$ scaling curves extended to TriviaQA and PopQA. 
Figure~\ref{fig:app_passk_all} illustrates the zero-shot pass@$k$ performance (up to $k=256$) across all evaluated benchmarks and model architectures.

The extended results confirm that the probability mass redistribution observed on NQ is a universal phenomenon. 
Across all datasets, post-RL models consistently exhibit a substantial advantage at low and medium sampling budgets ($k \le 64$), demonstrating that correct answers become significantly easier to elicit stochastically. 
Furthermore, as the sampling budget approaches $k=256$, the performance gap between pre-RL and post-RL models consistently narrows across all benchmarks. 
This universal convergence at high sampling budgets strongly reinforces our core conclusion: RL does not primarily inject novel facts, but rather systematically shifts deeply suppressed knowledge from the low-probability tail into highly accessible, lower-budget recall regimes.

\section{Extended Reward Dynamics of Suppressed Knowledge}
\label{apd:reward_dynamics}

\begin{figure}
    \centering
    \includegraphics[width=\linewidth]{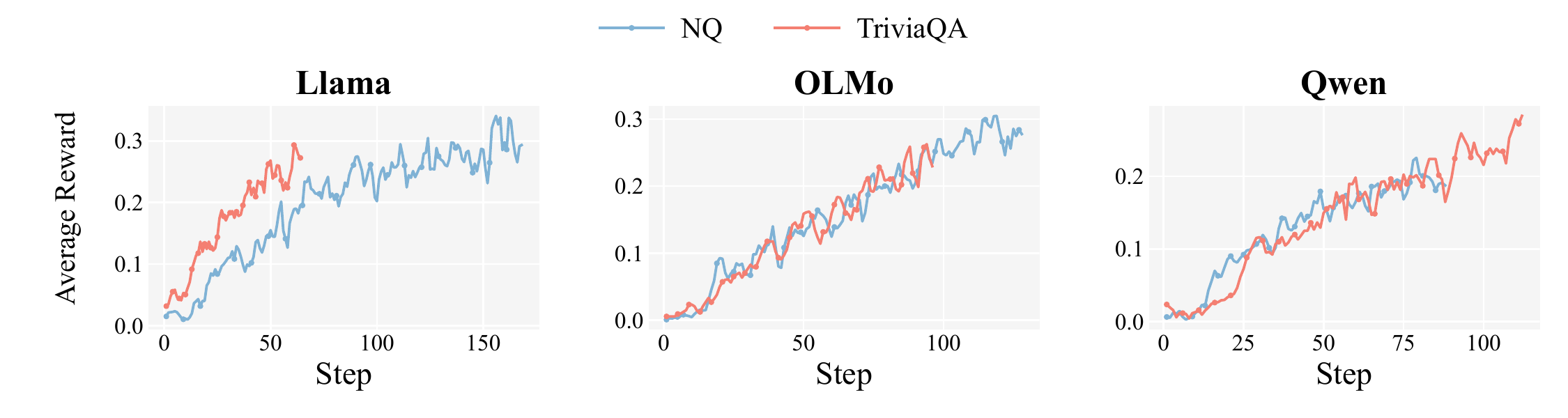}
    \caption{Training reward dynamics on the inaccessible@128 subset.}
    \label{fig:app_reward_dynamics_all}
\end{figure}

In the main text, we analyzed the training reward dynamics of the most challenging subset, i.e., examples with a $0/128$ pre-RL success rate, using Qwen as a representative case. 
In this section, we provide the complete reward dynamics across all three evaluated model families: OLMo, Llama, and Qwen. 
Figure~\ref{fig:app_reward_dynamics_all} tracks the average reward evolution for these inaccessible@128 examples throughout the RL training process.

The extended results reveal a highly consistent pattern across different model architectures. 
For all three models, the average reward starts near zero, confirming the extreme sparsity of correct responses under finite initial sampling. 
However, as training progresses, the reward curves exhibit a steady and continuous upward trajectory across all models and datasets.
This universal trend solidifies our core insight: finite-sample inaccessibility does not equate to the absence of knowledge. 
Regardless of the base model's architecture, continuous on-policy exploration during RL rollouts eventually samples some of these deeply suppressed facts. 
Once successfully generated, the corresponding reward signals are effectively captured and amplified, driving the remarkable recall improvements observed in our data attribution study. 

\renewcommand{\fcolorbox}[4][]{#4}
\begin{figure*}[t]
    \begin{tcolorbox}[
        right=5pt, left=5pt, top=5pt, bottom=5pt,
        toptitle=1mm, bottomtitle=1mm,
        colback=white,
        coltitle=white,
        colbacktitle=matisse,
        colframe=matisse,
        title=Prompt for Training-Test Deduplication, center title]
    \begin{minted}[fontsize=\small,autogobble,numberblanklines=false,breaklines]{markdown}
Task: Evaluate whether the core factual knowledge tested in the [Test Question] is semantically identical to the knowledge tested in the [Train Question].

Judgment Criteria (CRITICAL):
Think of each question as querying a Knowledge Graph triplet: (Subject Entity, Relation, Target Answer).
1. STRICT EQUIVALENCE (true): The questions ask for the EXACT SAME relation about the EXACT SAME subject entity. They are merely linguistic paraphrases of each other.
2. DIFFERENT RELATION (false): The questions share the same subject entity, but ask about different attributes or relations (e.g., "birth date" vs. "birth place").
3. DIFFERENT ENTITY / GENERALIZATION (false): The questions share the exact same answer, but they inquire about DIFFERENT subject entities. This represents compositional generalization, NOT contamination. (e.g., asking about the author of "Harry Potter" vs. the creator of "Hermione Granger").

---
### Few-Shot Demonstrations ###

Example 1:
[Test Question]: In what year did the RMS Titanic sink? (Answer: 1912)
[Train Question]: When did the Titanic go down? (Answer: 1912)
Output:
{
  "is_contained": true,
  "reason": "Both questions inquire about the exact same fact: (Titanic, sink_year, 1912). They are linguistic paraphrases."
}

...

Example 3:
[Test Question]: Who wrote the Harry Potter series? (Answer: J.K. Rowling)
[Train Question]: Which author created the character Hermione Granger? (Answer: J.K. Rowling)
Output:
{
  "is_contained": false,
  "reason": "They share the same answer, but target different subject entities ('Harry Potter series' vs. 'Hermione Granger'). This is factual generalization, not semantic overlap."
}

---
### Current Evaluation ###

[Test Question]: {test_question} (Answer: {test_answer})
[Train Question]: {train_question} (Answer: {train_answer})

Please strictly output the result in the defined JSON format:
    \end{minted}
    \end{tcolorbox}
    \caption{Prompt for LLM-based deduplication between training and test splits. The model determines whether a test QA pair and a candidate training QA pair target the same fact.}
    \label{fig:prompt_dedup}
\end{figure*}

\renewcommand{\fcolorbox}[4][]{#4}
\begin{figure*}[t]
    \begin{tcolorbox}[
        right=5pt, left=5pt, top=5pt, bottom=5pt,
        toptitle=1mm, bottomtitle=1mm,
        colback=white,
        coltitle=white,
        colbacktitle=matisse,
        colframe=matisse,
        title=Prompt for LLM-as-a-Judge, center title]
    \begin{minted}[fontsize=\small,autogobble,numberblanklines=false,breaklines]{markdown}
Your job is to look at a question, a gold target, and a predicted answer, and then assign a grade of either ["CORRECT", "INCORRECT"].

The following are examples of CORRECT predicted answers.
```
Question: What are the names of Barack Obama's children?
Gold target: Malia Obama and Sasha Obama
Predicted answer 1: sasha and malia obama
Predicted answer 2: Malia and Sasha Obama are the names of Barack Obama's children.
```
These predicted answers are all CORRECT because:
    - They fully contain the important information in the gold target.
    - They do not contain any information that contradicts the gold target.

The following are examples of INCORRECT predicted answers.
```
Question: What are the names of Barack Obama's children?
Gold target: Malia and Sasha
Predicted answer 1: Malia.
Predicted answer 2: Malia, Sasha, and Susan.
Predicted answer 3: Malia and Sasha, Malia and Sasha, Malia and Sasha, Malia and Sasha (repeated answer)
```
These predicted answers are all INCORRECT because:
    - A factual statement in the answer contradicts the gold target or contain repeated answer.


Here is a sample. Simply reply with either CORRECT or INCORRECT.

```
Question: {question}
Gold target: {target}
Predicted answer: {predicted_answer}
```

According to the gold target, please grade the predicted answer of this question as one of:
1.0: CORRECT
0.0: INCORRECT

Just return the numerical scores 1.0 or 0.0, with no text around it.
    \end{minted}
    \end{tcolorbox}
    \caption{The complete prompt used for LLM-as-a-Judge evaluation. Given a question, a gold target, and a predicted answer, the model returns a binary score of 1.0 (correct) or 0.0 (incorrect) based on semantic equivalence with the gold target.}
    \label{fig:prompt_judge}
\end{figure*}

\clearpage

\end{document}